\documentclass[12pt,journal,compsoc,onecolumn,draftclsnofoot]{IEEEtran}

\usepackage{scalerel}
\usepackage{tikz}
\usetikzlibrary{svg.path}

\definecolor{orcidlogocol}{HTML}{A6CE39}
\tikzset{
 orcidlogo/.pic={
 \fill[orcidlogocol] svg{M256,128c0,70.7-57.3,128-128,128C57.3,256,0,198.7,0,128C0,57.3,57.3,0,128,0C198.7,0,256,57.3,256,128z};
 \fill[white] svg{M86.3,186.2H70.9V79.1h15.4v48.4V186.2z}
 svg{M108.9,79.1h41.6c39.6,0,57,28.3,57,53.6c0,27.5-21.5,53.6-56.8,53.6h-41.8V79.1z M124.3,172.4h24.5c34.9,0,42.9-26.5,42.9-39.7c0-21.5-13.7-39.7-43.7-39.7h-23.7V172.4z}
 svg{M88.7,56.8c0,5.5-4.5,10.1-10.1,10.1c-5.6,0-10.1-4.6-10.1-10.1c0-5.6,4.5-10.1,10.1-10.1C84.2,46.7,88.7,51.3,88.7,56.8z};
 }
}

\newcommand\orcidicon[1]{\href{https://orcid.org/#1}{\mbox{\scalerel*{
\begin{tikzpicture}[yscale=-1,transform shape]
\pic{orcidlogo};
\end{tikzpicture}
}{|}}}}

\usepackage[T1]{fontenc}
\usepackage[retain-explicit-plus]{siunitx}
\usepackage{xcolor}
\usepackage{booktabs}
\usepackage{tabu}
\usepackage{tabularx}
\usepackage{tabulary}
\usepackage{multirow,bigdelim}
\usepackage{rotating}
\usepackage{makecell}
\usepackage{tablefootnote}
\usepackage{threeparttable}
\usepackage[ruled,vlined]{algorithm2e}
\usepackage{listings}
\usepackage{enumitem}
\usepackage[hyphens]{url}
\usepackage{lineno,hyperref}
\usepackage{comment}
\usepackage{tablefootnote}
\usepackage{enumitem,amssymb}
\usepackage{subcaption}
\usepackage{graphicx}
\usepackage{amssymb}
\usepackage{pifont}
\newcommand{\cmark}{\ding{51}}
\newcommand{\xmark}{\ding{55}}

\newlist{todolist}{itemize}{2}
\setlist[todolist]{label=$\square$}

\usepackage{blindtext}
\usepackage{hyperref}
\usepackage{nameref}

\newcounter{mylabelcounter}

\makeatletter
\newcommand{\labelText}[2]{%
\refstepcounter{mylabelcounter}%
\immediate\write\@auxout{%
 \string\newlabel{#2}{{\unexpanded{#1}}{\thepage}{{\unexpanded{#1}}}{mylabelcounter.\number\value{mylabelcounter}}{}}%
}%
}
\makeatother

\begin{document}

© 2024 IEEE. Personal use of this material is permitted. Permission from IEEE must be obtained for all other uses, in any current or future media, including reprinting/republishing this material for advertising or promotional purposes, creating new collective works, for resale or redistribution to servers or lists, or reuse of any copyrighted component of this work in other works. This version of the article has been accepted for publication, after peer review but is not the Version of Record which is available online at: \url{https://doi.org/10.1109/MIE.2023.3284203}.

\newpage

\title{Automatic generation of insights from workers' actions in industrial workflows with explainable Machine Learning}

\author{Francisco de Arriba-P\'erez\orcidicon{0000-0002-1140-679X}, Silvia García-Méndez\orcidicon{0000-0003-0533-1303}, Javier Otero-Mosquera\orcidicon{0000-0002-8552-5871}, Francisco J. González-Castaño\orcidicon{0000-0001-5225-8378}, and Felipe Gil-Castiñeira\orcidicon{0000-0002-5164-0855}
\IEEEcompsocitemizethanks{\IEEEcompsocthanksitem Francisco de Arriba-Pérez, Silvia García-Méndez, Javier Otero-Mosquera, and Francisco J. González-Castaño are with the Information Technologies Group, atlanTTic, University of Vigo, Vigo, Spain.
\\E-mail: sgarcia@gti.uvigo.es
\IEEEcompsocthanksitem Felipe Gil-Castiñeira is with Ancora Mobile S.L., Area Portuaria de Bouzas, Edificio Consorcio Zona Franca de Vigo, Oficina C2, Vigo, Spain.}}

\markboth{IEEE Industrial Electronics Magazine}
{de Arriba-Pérez \MakeLowercase{\textit{et al.}}: Automatic detection of worker \textsc{kpi}s in an industrial scenario}

\makeatletter
\long\def\@IEEEtitleabstractindextextbox#1{\parbox{0.922\textwidth}{#1}}
\makeatother

\IEEEtitleabstractindextext{%
\begin{abstract}
New technologies such as Machine Learning (\textsc{ml}) gave great potential for evaluating industry workflows and automatically generating key performance indicators (\textsc{kpi}s). However, despite established standards for measuring the efficiency of industrial machinery, there is no precise equivalent for workers' productivity, which would be highly desirable given the lack of a skilled workforce for the next generation of industry workflows. Therefore, an \textsc{ml} solution combining data from manufacturing processes and workers' performance for that goal is required. Additionally, in recent times intense effort has been devoted to explainable \textsc{ml} approaches that can automatically explain their decisions to a human operator, thus increasing their trustworthiness. We propose to apply explainable \textsc{ml} solutions to differentiate between expert and inexpert workers in industrial workflows, which we validate at a quality assessment industrial workstation. Regarding the methodology used, input data are captured by a manufacturing machine and stored in a \textsc{n}o\textsc{sql} database. Data are processed to engineer features used in automatic classification and to compute workers' \textsc{kpi}s to predict their level of expertise (with all classification metrics exceeding \SI{90}{\percent}). These \textsc{kpi}s, and the relevant features in the decisions are textually explained by natural language expansion on an explainability dashboard. These automatic explanations made it possible to infer knowledge from expert workers for inexpert workers. The latter illustrates the interest of research in self-explainable \textsc{ml} for automatically generating insights to improve productivity in industrial workflows.
\end{abstract}

}

\maketitle

\IEEEdisplaynontitleabstractindextext

\IEEEpeerreviewmaketitle


\section{Introduction}

As indicated by Deloitte in its \textit{Creating pathways for tomorrow’s workforce today} report\footnote{Available at \url{https://www2.deloitte.com/us/en/insights/industry/manufacturing/manufacturing-industry-diversity.html}, June 2023.}, finding qualified workers with specific skills is nowadays 1.4 times harder than four years ago. Due to this skill gap, over 2.1 million positions in the United States will be open by 2030. The latter will exacerbate the effect in the higher levels of an organization of productivity at lower levels \cite{Lather2019}, as well as the problem of workers turnover in manufacturing companies, motivating these companies to train their workers themselves and cost-effectively analyze workers' skills \cite{Patalas-Maliszewska2020}.

Large-scale industrial data analytics has been addressed relatively recently \cite{Cheng2018}. Companies collect information from their machinery and their production environments \cite{Saqlain2019}, the movements of their workers, and their interactions throughout the workflows \cite{Bavaresco2021}, supported by sensors, interconnected wireless devices, and positioning technologies. In general, companies, thus, have started to realize the potential of new technologies such as Artificial Intelligence (\textsc{ai}) and Machine Learning (\textsc{ml}) to generate key performance indicators \cite{Ramachandran2022} and increase productivity through automation \cite{Knights2021}. Of particular interest to this work is using mobile assistants to guide the workers along these workflows \cite{Tao2019}. Knowledge can be mined from this information in favor of process and productivity enhancement \cite{Rauch2020}. 

In recent times intense effort has been devoted to explainable \textsc{ml} approaches that can automatically describe their decisions to make them understandable to a human operator, thus, increasing their trustworthiness \cite{Gunning2019}. In this paper, we propose to apply explainable \textsc{ml} solutions to differentiate between expert and inexpert workers in industrial workflows. From the resulting automatic explanations of classifier decisions, we demonstrate in a case study that it is possible to infer helpful knowledge that can be provided to inexpert workers. As far as we know, this is an entirely novel approach to automatic knowledge extraction from manufacturing workflows.

The rest of this paper is organized as follows. Section \ref{sec:related_work} discusses the background in automatic detection of key performance indicators (\textsc{kpi}s) by paying particular attention to the analysis of worker performance in industrial scenarios. Section \ref{sec:proposed_method} describes the proposed architecture for explainable skill level classification. Section \ref{sec:experimental_results} presents the experimental testbed of a representative case study and the results obtained in classifying skill level and automatic inference of knowledge to improve worker performance. Finally, Section \ref{sec:conclusions} concludes the paper and proposes future research.

\section{Related work}
\label{sec:related_work}

Industry 4.0 seeks to create a fully connected environment in which the workers assume a crucial role (\textit{i.e.}, the human-centrality concept of Worker 4.0 or Operator 4.0 \cite{Kaasinen2020}) \cite{Bian2021}. Worker-machine interactions can be collected with wearables and distributed sensors, enabling the analysis of these interactions to improve occupational safety and health and to create a flexible and configurable productivity system \cite{Patel2022}. Consequently, management platforms gather workers' data from which valuable insights for optimizing labor can be obtained.

The latter is related to and builds upon the Industrial Internet of Things (\textsc{ii}o\textsc{t}), one of the pillars of Industry 4.0 supporting the direct exchange of information between intelligent objects, people, and processes \cite{Jasperneite2020}. In this paradigm, in an industrial context, industrial sensors, actuators, and elements integrating them, such as mobile robots, become communication nodes \cite{Grau2021}.

Another critical field for Industry 4.0, mainly Worker/Operator 4.0, is \textsc{ai}. \textsc{ml} algorithms learn and evolve automatically from data sets that conform to the knowledge base of domain-specific problems, like in analyzing bio-metric signals and detecting patterns in economic environments. In the industrial domain, predictive maintenance \cite{Loisel2021}, prediction of service outages \cite{Konstantinou2020}, and analysis of workers' and robots' efficiency \cite{Lee2017,Bhattacharya2020} are representative examples.

However, most existing research on Industry 4.0 intelligence focuses on machines owing to the extensive automation of manufacturing plants nowadays, rather than on the characterization of the efficiency and productivity of workers connected within the industrial environment. For example, in Hernandez \textit{et al.} \cite{HernandezBennetts2019}, a heterogeneous measurement system composed of static detection nodes and continuous sampling is proposed, complemented by localized measurements collected during sporadic detection by a mobile robot. In Segreto and Teti \cite{Segreto2019}, a novel decision-making solution for minimizing downtime in machines is presented based on neural networks that use data from multiple sensors to obtain an exact classification of the status and performance of the manufacturing process.
 
Currently, the efficiency of industrial equipment is measured with the average Overall Equipment Effectiveness (\textsc{oee}) indicator \cite{Braglia2021}. This \textsc{kpi} was first used in Japan and was quickly accepted and recognized as an international standard by the leading industrial players \cite{Muthalib2020}. It combines the efficiency of the machines in a single measure and aspects such as time of use, number of objects or actions correctly executed, and production rate. Nevertheless, regardless of the significance of the operational skills of workers to the quality of processes and the competitive advantage of organizations \cite{Fallucchi2020}, there is no precise equivalent metric for the efficiency of workers. In some environments, a direct adaptation of \textsc{oee}, such as Overall Labor Effectiveness (\textsc{ole}), is used as a \textsc{kpi} \cite{Hernandez2021}. It analyzes the level of productivity based on the time dedicated to a task and the quantity and quality of the products obtained. However, it has a highly subjective component, and depending on the industrial or business target, it cannot be easy to quantify.

Alternative approaches explore workers' performance through dashboards and surveys so that supervisors can follow, monitor, and analyze the execution of tasks to organize their teams \cite{Pradhan2017} efficiently. Other works address new efficiency measurements more focused on obtaining Objectives and Key Results (\textsc{okr}) to estimate the degree of goal achievement in working environments \cite{Zhou2018}.

A highly related variable to worker efficiency is the location in the workspace. For example, Sun \textit{et al.} \cite{Sun2021} proposed to use Bluetooth Low Energy (\textsc{ble}) beacons and a positioning model based on convolutional neural networks (\textsc{cnn}) for worker location, achieving precisions around \SI{98}{\percent} in work environments. Finer capture of worker movements is challenging. Amorin \textit{et al.} \cite{Amorim2021} identified with \textsc{ml} techniques operator gestures and behaviors from 3D body sensors.

Given the machine-human \textsc{kpi} gap mentioned above, and more closely related to our work, Forkan \textit{et al.} \cite{Forkan2019} presented a novel \textsc{ii}o\textsc{t} solution to monitor, evaluate and improve productivity through \textsc{kpi}s based on the recognition of worker activity with a distributed platform composed of wearable sensors. Lather \textit{et al.} \cite{Lather2019} assessed worker performance with an \textsc{ml} model based on variables collected by hand, using individual (gender, age, education, etc.), domain-specific (industrial sector, years of experience, etc.) and socioeconomic (income, area of location, etc.) questionnaires. Al \textit{et al.} \cite{AlJassmi2019} considered the effect of physiological signals collected with portable sensors. They concluded that there is a moderate positive correlation between the emotional state of the workers and their productivity. Davoudi \textit{et al.} \cite{DavoudiKakhki2019} presented an approach for incident prediction in industrial environments with diverse supervised \textsc{ml} classification algorithms to relate severity ranges, age of workers, and seniority in the company, exceeding \SI{90}{\percent} accuracy. Finally, some works on automatic performance evaluation have considered the level of expertise of the workers. Fantini \textit{et al.} \cite{Fantini2020} created a methodology for the design and evaluation of workflows by considering differential worker features (\textit{i.e.}, abilities, knowledge, and skills). They applied their theoretical framework in two industrial use cases, aerospace and compressor plants. However, their approach was not validated at execution time, only by design. Peruzzini \textit{et al.} \cite{Peruzzini2020} also proposed a theoretical framework for physical ergonomics and operators' mental workload evaluation using wearable sensors and eye-tracking protocols. Their solution was deployed with virtual prototypes that workers used to interact with the digital twin of agriculture and industrial vehicle manufacturing plants. No results on worker performance detection or improvement were provided. Patalas-Maliszewska and Halikowsk \cite{Patalas-Maliszewska2020} analyzed the workflow proficiency of new employees in a manufacturing company using deep learning from video data. Belkadi \textit{et al.} \cite{Belkadi2020} proposed a Proper Generalized Decomposition (\textsc{pgd}) model to assist the workers according to their assigned tasks, given their skills.

However, none of the works in this review describe the reasons behind their predictions, thus, lacking transparency from the perspective of the end users, who must unthinkingly trust the outcomes of the automatic decision systems, unlike those of traditional intelligible manual approaches. Note that these explanations could also enhance industrial productivity by suggesting the application of corrective measures. In fact, the lack of explainability represents one of the main obstacles to adopting \textsc{ml} techniques in many sectors of economic activity \cite{Carletti2019,Peres2020,Crawford2021}. There exists more prior work in this regard. The work by Carletti \textit{et al.} \cite{Carletti2019} is one of the few exceptions. The authors addressed the lack of interpretability of \textsc{ml} models with a feature importance explainability solution using the Isolation Forest model for anomaly detection, a field of great importance in industrial scenarios.

Regarding commercial solutions, a well-defined sector of technology-oriented companies is working on digital workflow optimization (Contextere\footnote{Available at \url{https://contextere.com}, June 2023.}, Evolaris\footnote{Available at \url{https://www.evolaris.net/en}, June 2023.}, Plataine\footnote{Available at \url{https://www.plataine.com}, June 2023.}, Workerbase\footnote{Available at \url{https://workerbase.com}, June 2023.}, etc.) and Worker/Operator 4.0 solutions (Ancora Worker Connect aka \textsc{awc}\footnote{Available at \url{https://www.ancoramobile.com}, June 2023.}, Datch\footnote{Available at \url{https://www.datch.io}, June 2023.}, HelixAI\footnote{Available at \url{https://www.askhelix.io}, June 2023.}, Intelligent Insights\footnote{Available at \url{https://4i.ai}, June 2023.}, \textsc{knowron}\footnote{Available at \url{https://www.knowron.com}, June 2023.}, Lucas Systems\footnote{Available at \url{https://www.lucasware.com}, June 2023.}, Screevo\footnote{Available at \url{https://screevo.ai}, June 2023.}, Simsoft Industry\footnote{Available at \url{https://www.simsoft-industry.com/en}, June 2023.}, etc.). However, most of these solutions focus almost exclusively on process management and set aside performance feedback (this is the case of Evolaris with their live support call functionality, HelixAI optimized access to data and information, and Intelligent Insights, whose main objective is to understand customer behavior), hence the interest of further research in the field of artificial intelligence combining data from the manufacturing process with efficiency logs of the workers.

\subsection{Research contribution}

This work contributes with a novel solution for automatically assessing worker \textsc{kpi}s in an industrial workflow scenario. The system can differentiate workers by their skill level and explain its decisions. It is shown that automatic explainability can be exploited to automatically transfer knowledge from expert to inexpert workers, in line with the need for tools for the in-house formation of a qualified workforce in the industry of the future.

Table \ref{tab:comparison} summarizes some relevant works from the literature on machinery and workers' performance profiling methods and their automatic explainability capabilities (\textit{i.e.}, we exclude works on other problems, such as anomaly detection, which lies out of the scope of this paper). Observe that sensor data is the most specific information to assess machinery and worker performance. Summing up, the differential nature of our proposal is that it not only assesses worker competence by combining supervised \textsc{ml} and \textsc{kpi}s but also automatically explains its decisions.

\begin{table*}[!htbp]
\centering
\caption{\label{tab:comparison}Comparison with previous works.}
\begin{tabular}{p{3cm}ccp{4.5cm}c}
\toprule
\multirow{1}{*}{\textbf{Authorship}} & \textbf{Profiling target} & \textbf{Method} & \textbf{Input data} & \textbf{Explainability} \\
\midrule

Hernandez \textit{et al.} \cite{HernandezBennetts2019} & Machinery & Simulation & Sensor data & \multirow{2}{*}{\xmark}\\

Segreto and Teti \cite{Segreto2019} & Machinery & Deep learning & Sensor data\\ \midrule

\multirow{3}{*}{Lather \textit{et al.} \cite{Lather2019}} & \multirow{3}{*}{Worker} & \multirow{3}{*}{Supervised \textsc{ml}} & User (age, etc.), domain (sector, experience, etc.), social-economic features & \multirow{7}{*}{\xmark}\\

Forkan \textit{et al.} \cite{Forkan2019} & Worker & Supervised \textsc{ml} + \textsc{kpi}s & Motion sensor data \\

Al \textit{et al.} \cite{AlJassmi2019} & Worker & Statistical analysis & Physiological sensor data &\\ 

Patalas-Maliszewska and Halikowsk \cite{Patalas-Maliszewska2020} & \multirow{2}{*}{Worker} & \multirow{2}{*}{Deep learning} & \multirow{2}{*}{Video data}\\

\midrule\midrule

\textbf{Our proposal} & Worker & Supervised \textsc{ml} + \textsc{kpi}s & Manufacturing sensor data & \cmark\\
\bottomrule
\end{tabular}
\end{table*}

\section{Methodology: explainable Machine Learning \& \textsc{kpi}s }
\label{sec:proposed_method}

\begin{figure*}[!htbp]
\centering
\includegraphics[scale=0.15]{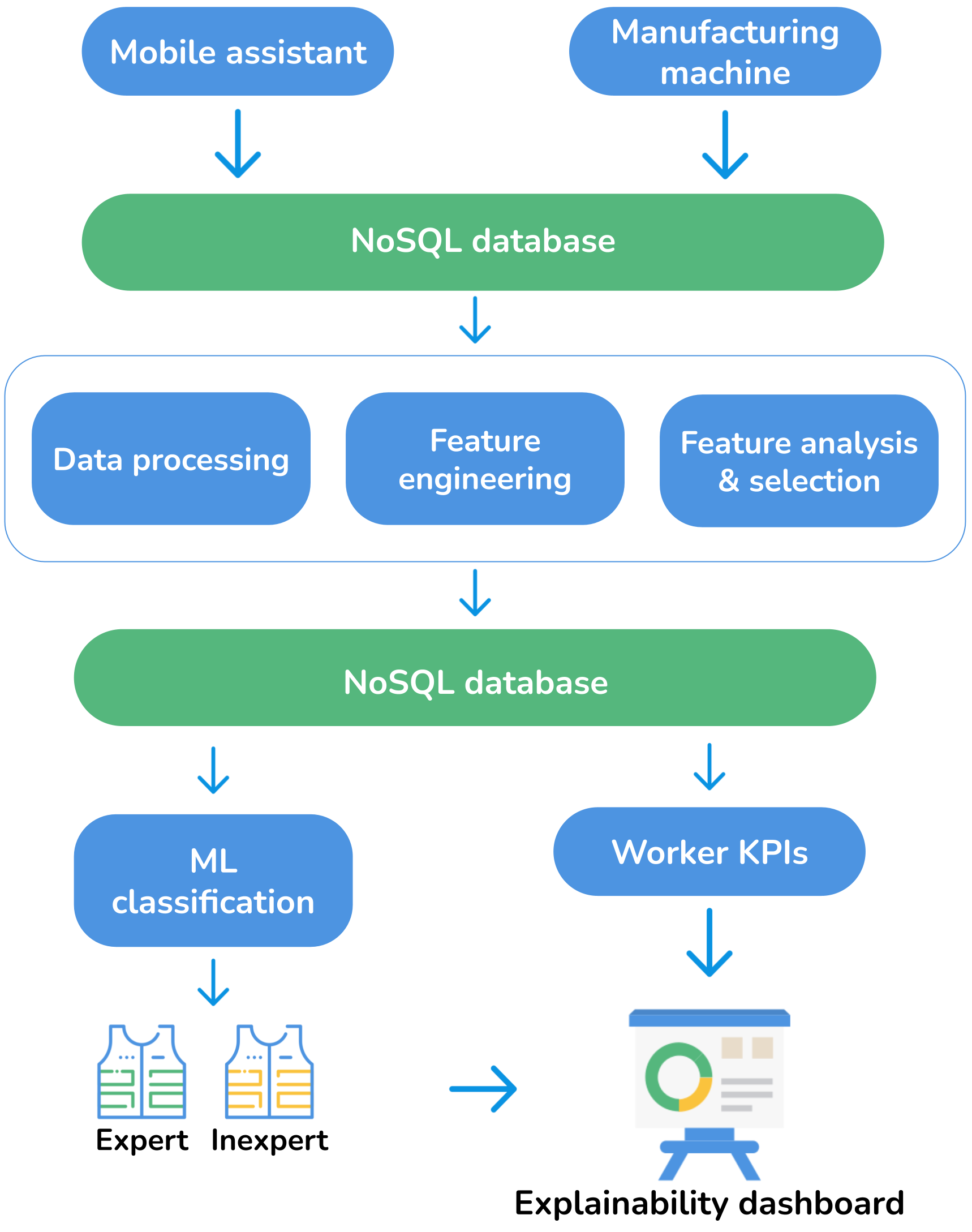}
\caption{\label{fig:scheme}Scheme of the solution.}
\end{figure*}

Figure \ref{fig:scheme} shows the scheme of the proposed architecture. Input data are captured by a manufacturing machine (Section \ref{sec:input_data}). A No\textsc{sql} database stores selected data for subsequent processing (Section \ref{sec:nosqldatabase}). Note that this type of database is often used by Manufacturing Execution Systems (\textsc{mes}s) \cite{Kozjek2017,Lee2018,Cui2020}, within which workflow management can be integrated. Data processing, feature engineering, analysis, and selection procedures (Section \ref{sec:data_processing}) are performed before calculating worker \textsc{kpi}s (Section \ref{sec:worker_kpis}) and classifying them by expertise level (classification stage in Section \ref{sec:classification}). In the end, the predictions on the expertise level of the workers, along with the corresponding automatic insights and \textsc{kpi}s are textually and visually described on the explainability dashboard (Section \ref{sec:explainability}).

\subsection{Industrial workflow: manufacturing machines \& mobile assistant}
\label{sec:input_data}

\begin{figure*}[!htbp]
\centering
\includegraphics[scale=0.15]{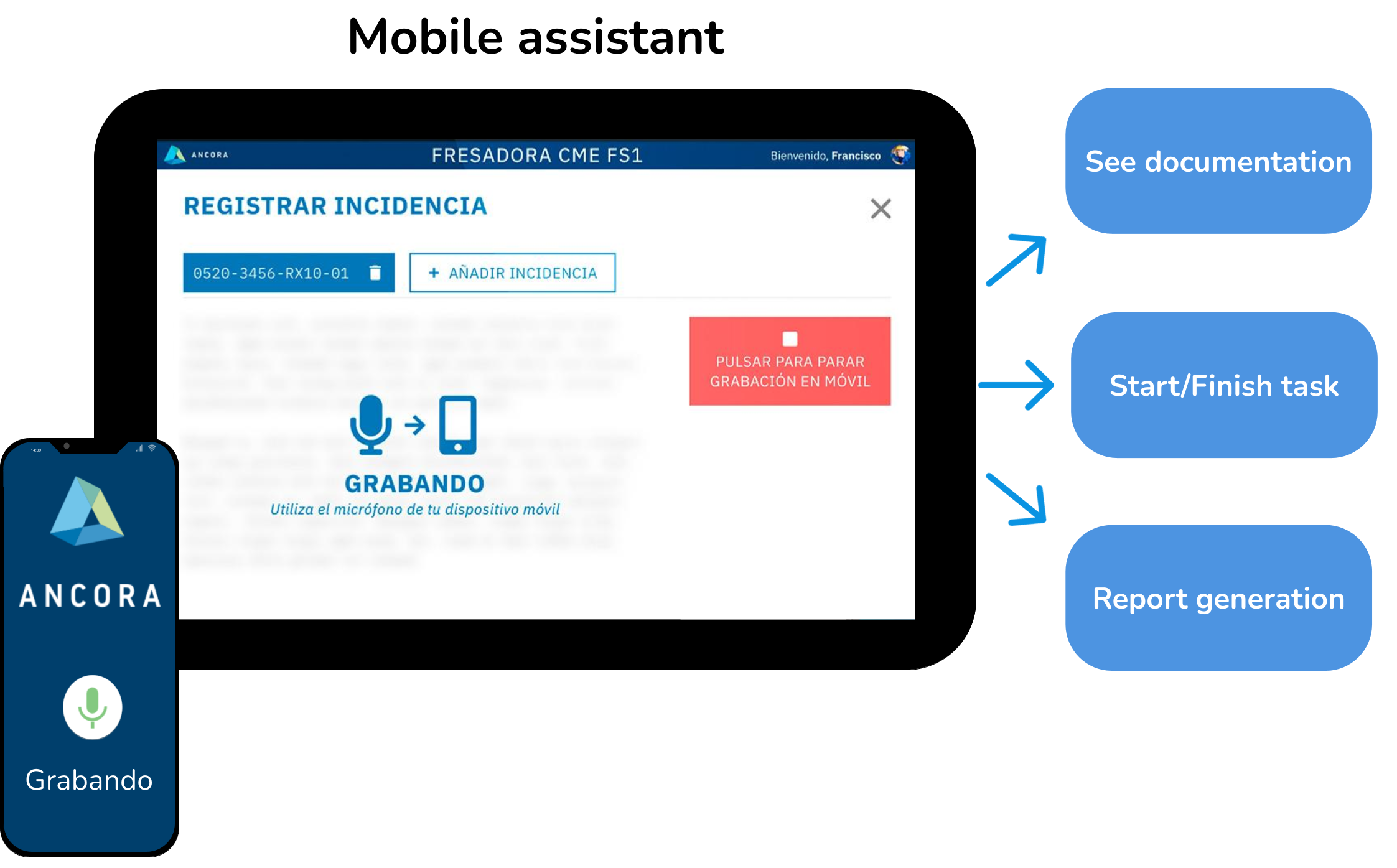}
\caption{\label{fig:awc}Mobile assistant.}
\end{figure*}

The mobile assistant guides the workers through the industrial workflow composed of different tasks at manufacturing machines and allows them to check task starts and terminations. More in detail, the operators use \textsc{awc} (see Figure \ref{fig:awc}) as the task manager assistant. It allows them to consult the documentation, report incidences, indicate the start and end of the manufacturing task, etc., using voice commands so it does not interfere with the tasks. The exact process performed by the operators is as follows: (\textit{i}) they start the assistant to consult the documentation, then (\textit{ii}) they select the task, and the task starts; (\textit{iii}) when the pieces are correct or when the number of attempts has been exceeded, the task finishes. During the task, using the same mobile assistant, the workers can report incidents (\textit{e.g.}, stuck parts, unexpected stops of the machine, etc.).

The manufacturing machine gathers data through different sensors, such as weight sensors and artificial vision. These sensors detect the number of pieces stored and reloaded; and the validity of the pieces. A \textsc{json} is sent to the NoSQL database for each piece finished.

\subsection{Non-relational database}
\label{sec:nosqldatabase}

\begin{figure*}[!htbp]
\centering
\includegraphics[scale=0.15]{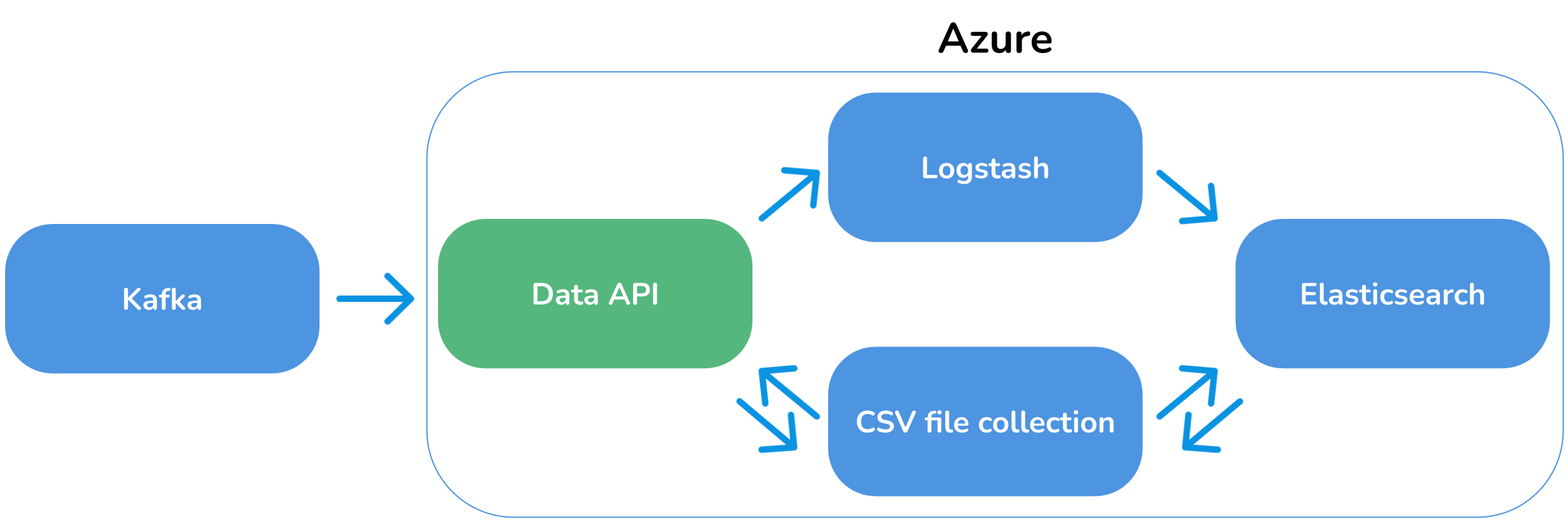}
\caption{\label{fig:elasticsearch}Storage flow.}
\end{figure*}

The database to store the logs from \textsc{awc} and the manufacturing machine is a non-relational NoSQL database in the Azure\footnote{Available at \url{https://azure.microsoft.com/es-es}, June 2023.} ecosystem (see Figure \ref{fig:elasticsearch}). Permanent information channels or connectors were implemented using Kafka\footnote{Available at \url{https://kafka.apache.org}, June 2023.} and Data API Azure methods. These connectors establish communication routes under subscription, which allows efficient management of communication queues and, consequently, proper solution scaling.

The NoSQL database uses the Elasticsearch engine\footnote{Available at \url{https://www.elastic.co}, June 2023.}. A non-relational paradigm is adequate for our scenario because numerical, machine-dependent features and non-numerical data from the mobile assistant resulting from human interactions are often intrinsically unstructured. Homogeneous \textsc{json} templates are managed by the Logstash module\footnote{Available at \url{https://www.elastic.co/es/logstash}, June 2023.}, which collects data from the \textsc{json} and adapts them to queries for storage in Elasticsearch. An additional module was designed to collect data from Elasticsearch in a specific time interval and save these data in a \textsc{csv} file with all the necessary information to perform the experiments. 

\subsection{Data processing, feature engineering, analysis \& selection}
\label{sec:data_processing}

This stage adapts input data into a unified format due to the variety of information gathered at the workstation. 

Feature engineering seeks to profile manufactured pieces and worker actions. For the characterization of a piece, temporary data and whether the piece is invalid or valid are registered. For the characterization of the actions, features comprise temporary data, incidences in the workstation, and other information on the pieces; materials' reloads in the machines; and interactions via the mobile assistant.
 
Feature selection is based on two approaches: Pearson correlation analysis and selection using a meta-transformer wrapper over a tree-based \textsc{ml} model. The selected features are those that are chosen with both methods.

For the first approach, we apply the Pearson correlation coefficient (\ref{pearson}) \cite{Benesty2009}, where -1 and 1 extremes indicate that two variables have the strongest possible inverse and direct relations, respectively. Features are considered irrelevant if the absolute value of their correlation with the target variable is less than a configurable threshold $\delta$.

\begin{equation}
r_{xy} = \frac{\sum (x_i - \overline{x}) (y_i - \overline{y})}{\sqrt{\sum (x_i - \overline{x})^2} \sqrt{\sum (y_i - \overline{y})^2}}
\label{pearson}
\end{equation}

Instead of using other feature selection methods, such as those based on percentile or variance, the second approach is model-agnostic \cite{Burkart2021}, ensuring that only the most relevant features enter the classification stage. 

Specifically, it is based on a meta-transformer wrapper over a tree-based \textsc{ml} model, \textit{i.e.}, by exploiting importance weights to detect and discard impurity-based features. More in detail, this feature selection method is based on the Mean Decrease in Impurity (\textsc{mdi}) method \cite{Breiman2017} that computes the average proportion of sample splits by the feature in every node over all trees of the tree-based model. The features with a \textsc{mdi} lower than the \textsc{mdi} average are discarded.

\subsection{Worker \textsc{kpi}s}
\label{sec:worker_kpis}

Worker \textsc{kpi}s are computed from features related to pieces and worker actions. We chose them inspired by previous work with good results in industrial applications \cite{Forkan2019}. They measure intra- and inter-worker performance over time. Intra-worker patterns are intended to detect specific behaviors resulting from external factors. In contrast, inter-worker patterns compare a worker's behavior with the rest of the team. These \textsc{kpi}s are of interest to studying the evolution of inexpert and expert workers.

\subsection{Machine Learning classification}
\label{sec:classification}

 \textsc{ml} classification seeks to detect: (\textit{i}) the level of expertise of the workers in the workflow and (\textit{ii}) whether an inexpert or expert worker carries out a task.

The experimental plan is thus divided into two scenarios:

\begin{description}

\item \textbf{Scenario 1} analyzes the piece manufacturing process. The target feature identifies whether an inexpert or expert worker has produced the piece.

\item \textbf{Scenario 2} focuses on task performance. The worker must produce a number of pieces. The classification stage predicts whether an inexpert or expert worker has performed the task.

\end{description}

Note that we apply supervised approaches in both scenarios. Target labels are extracted from the experience and position in the human resources records of the company. The \textsc{ml} models used were selected based on their performance-time trade-off in typical problems \cite{Angelopoulos2019} and the promising results obtained by competing works in the literature \cite{Lather2019,Forkan2019}.

\begin{itemize}

 \item \textbf{Support Vector Classifier} (\textsc{svc}) \cite{Alam2020} is highly flexible in terms of error penalties and hyper-parameters. Note that the performance of the \textsc{svc} model depends on the kernel used (the most popular are linear, polynomial, Radial Basis Function \textit{rbf}, and sigmoid). Therefore, experimental results were computed with all these kernels. This model was selected since it reduces the generalization error upper bound instead of the training error \cite{Rodan2014}. Thus, it has the potential to provide accurate results (globally optimal) with a reduced number of training samples in a short period, which is rather appropriate in our industrial use case. It is also less prone to overfitting than other alternatives.
 
 \item \textbf{Random Forest classifier} (\textsc{rf}) \cite{Schonlau2020}, a meta estimator composed of several Decision Tree (\textsc{dt}) classifiers. It averages them seeking a trade-off between accuracy and over-fitting control. It computes as the mean squared error (\textsc{mse}) the distance between a node and the actual value to follow the most appropriate branch to obtain a classification decision from the options in the forest. Its most important advantage lies in extracting hierarchical and nonlinear relationships between the training and target features, which are expected to exist in our industrial data set, by exploiting an ensemble strategy. Note that the ensemble approach enhances the accuracy and the robustness of the predictions \cite{Everingham2016}.

 \item \textbf{AdaBoost classifier} (\textsc{ab}) \cite{Zizka2019}, also a meta-estimator. It uses several instances of a classifier (with multi-class support) to classify complex data sets. The particularity of this model lies in different adjustments of the feature weights in each instance. The overall prediction is computed from the weighted sum of the predictions of the instances. Accordingly, this ensemble model is especially advantageous in classification problems with highly unbalanced data sets common in industrial research. It also significantly reduces the prediction bias (\textit{i.e.}, overfitting) since boosting combinations decrease the variance \cite{martin2017}.

\end{itemize}

The evaluation metrics used were accuracy and macro and micro values for precision, recall, and \textit{F}-measure, along with the elapsed total time of the \textsc{ml} models. More in detail, the accuracy indicates the performance of the models considering the number of correct predictions. Precision refers to the quality of positive predictions, while recall indicates the model's ability to detect positives (the higher the recall, the more positives are detected). The \textit{F}-measure is a combined metric that considers both precision and recall to evaluate model effectiveness. Finally, regarding the difference between micro and macro values, the former computes the corresponding metric (precision, recall, or \textit{F}-measure in our case) independently for each class, while the latter aggregates the contributions of all classes to average the metric.

\subsection{Explainability}
\label{sec:explainability}

Natural language explanations about the workers' performance are derived from relevant features in the classification processes. These are the automatic insights that are extracted in the explainability stage. They are presented textually (as natural language templates) and visually (on the explainability dashboard). The information provided comprises:

\begin{itemize}
 \item The most relevant features of the \textsc{ml} model.

 \item Intra- and inter-worker performance analysis based on weekly and daily feature values, respectively.

\end{itemize}

The explainability module is based on Local Interpretable Model-agnostic Explanations (\textsc{lime}) \cite{Goode2021}. This approach is divided into three main stages: (\textit{i}) data simulation and transformation to fit the black-box model, (\textit{ii}) prediction using the black-box algorithm over the simulated data to obtain the explainer model, and (\textit{iii}) interpretation of the explainer model in terms of features involved and their relevance in the prediction.

\section{Experimental results}
\label{sec:experimental_results}

In this section, the experimental testbed and the data set are described (sections \ref{sec:experimental_dataset} and \ref{sec:input_data_results}, respectively). 
Section \ref{sec:data_processing_results} details the outcome of the data processing and feature engineering, analysis, and selection procedures. Section \ref{sec:classification_results} discusses the classification results for the two scenarios considered, including a performance comparison with related works from the literature. Section \ref{sec:worker_kpis_results} presents the proposed \textsc{kpi}s to support the explainability of the classification decisions. Finally, Section \ref{sec:explainability_results} presents the knowledge extracted from the explainability stage.

The Elasticsearch server is composed of a Kubernetes\footnote{Available at \url{https://kubernetes.io}, June 2023.} cluster with three standard\_B4ms instances with four \textsc{cpu} each and the following specifications:
\begin{itemize}
  \item \textbf{RAM}: \SI{16}{\giga\byte}
  \item \textbf{Disk}: \SI{16}{\giga\byte}
\end{itemize}

The tablet where \textsc{awc} is installed has the following specifications:
\begin{itemize}
  \item \textbf{Model}: Lenovo Tab M10 \textsc{hd} Gen 2
  \item \textbf{Operating System}: Android 10
  \item \textbf{Processor}: Octa-core (4x\SI{2.30}{\giga\hertz} Cortex-A53 \& 4x\SI{1.80}{\giga\hertz} Cortex-A53)
  \item \textbf{GPU}: PowerVR GE8320
  \item \textbf{RAM}: \SI{4}{\giga\byte}
  \item \textbf{Disk}: \SI{64}{\giga\byte}
\end{itemize}

All \textsc{ml} experiments were performed using a server with the following hardware specifications:
\begin{itemize}
 \item \textbf{Operating System}: Ubuntu 18.04.2 LTS 64 bits
 \item \textbf{Processor}: Intel\@Core i9-10900K \SI{2.80}{\giga\hertz}
 \item \textbf{RAM}: \SI{96}{\giga\byte} DDR4
 \item \textbf{Disk}: \SI{480}{\giga\byte} NVME + \SI{500}{\giga\byte} SSD
\end{itemize}

\subsection{Testbed elements}
\label{sec:experimental_dataset}

New technologies such as \textsc{ai} enable the efficient management of flexible workflows \cite{Hawash2020}. Even though human operators are still essential, they can be artificially augmented with aids such as collaboration robots (cobots). In contrast to industrial robots, cobots work side by side with the operators freeing them from repetitive and trivial tasks to attend to more complex and creative work, and they can be reprogrammed for different tasks throughout the life of the workflow in the production line \cite{Kadir2018}.

\begin{figure*}[!htbp]
\centering
\includegraphics[scale=0.15]{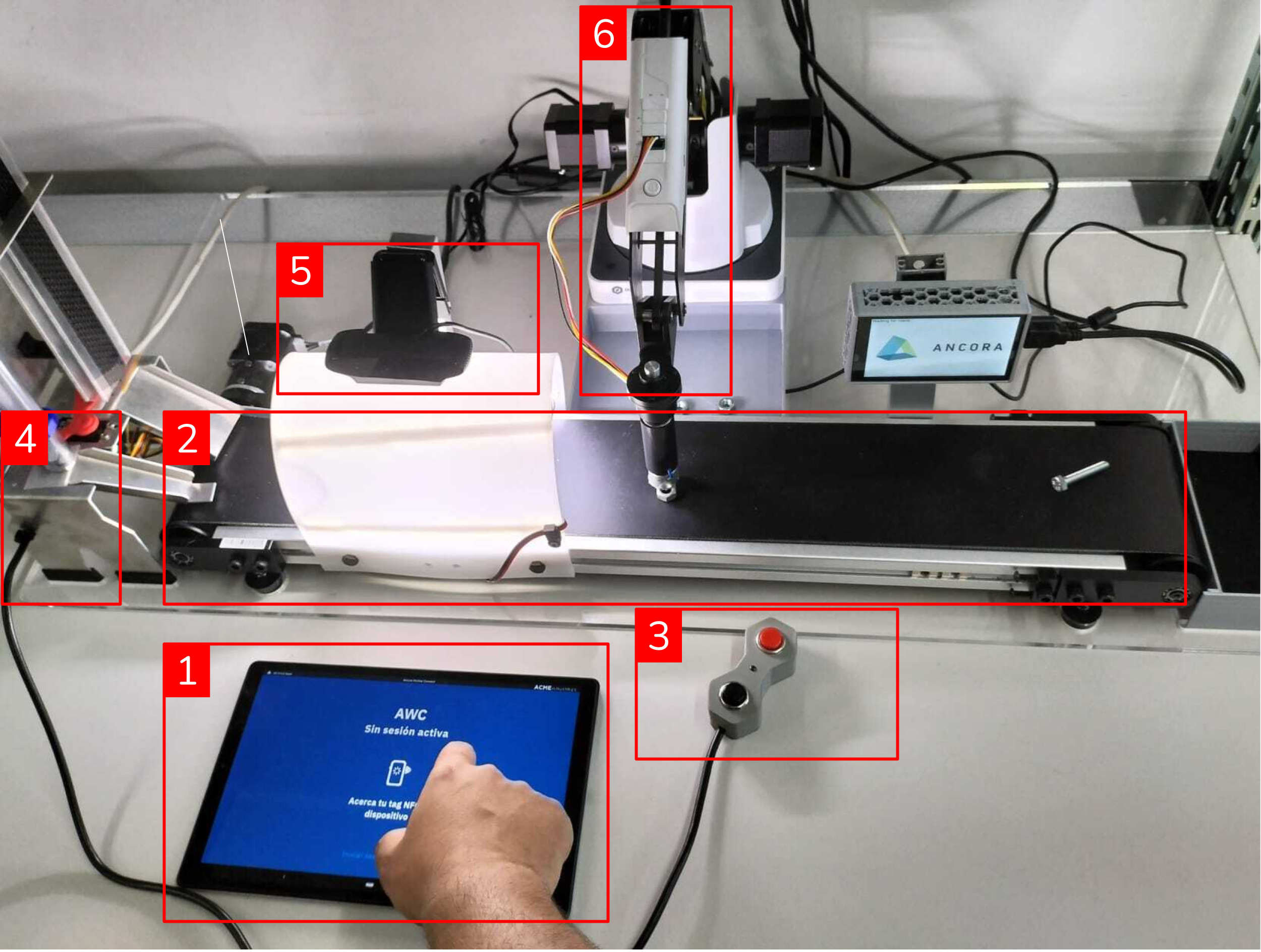}
\caption{\label{fig:testbed_awc}Testbed (1. Mobile assistant, 2. Conveyor belt, 3. Main switch, 4. Weight sensor, 5. Camera, 6. Cobot).}
\end{figure*}

For this reason, we have chosen a representative testbed with a cobot. Figure \ref{fig:testbed_awc} shows the cobot workstation and the mobile assistant screen (1). The purpose of the workstation is to perform a quality test of the pieces that are fed at the left (4) of the conveyor belt (2) by analyzing images of the pieces with a digital camera (5). The cobot arm (6) picks and retires the pieces whose dimensions are incorrect. The operator feeds the pieces into the system, which are taken from a warehouse tray (not shown). In the test, to emulate wrong pieces at will, the workstation was trained to recognize bots and nuts using the \texttt{Open\textsc{cv}} library\footnote{Available at \url{https://github.com/opencv/opencv}, June 2023.} with the \textsc{yolo} object detection module\footnote{Available at \url{https://opencv-tutorial.readthedocs.io/en/latest/yolo/yolo.html}, June 2023.}. Nuts emulated the wrong pieces (the cobot was retiring a nut when the picture was taken). Before starting a session, the worker consults the instructions on the mobile assistant screen (1), and then registers the session initiation on the mobile assistant (1), and executes actions until \num{7} valid pieces are obtained with a maximum of \num{12} attempts. At that moment, the worker registers that the session has been completed on the mobile assistant (1). After the worker takes the pieces from the warehouse tray to the left of the scene, there is a space for manipulations in a scaling buffer to the left (not shown).

\subsection{Data Set}
\label{sec:input_data_results}

The data set was collected from 5 workers (3 experts and 2 inexpert workers) during a week. Inexpert workers received previous training through the mobile assistant manual before the sessions.

Table \ref{tab:features_piece_task} details the features collected in the two scenarios related to piece and worker action profiling by the mobile assistant or the machine sensors depending on the case. All the features are numerical except for features \#5 and \#12, which are both Boolean and refer to piece type (true: valid, false: invalid).

\begin{table*}[!htbp]
\centering
\caption{\label{tab:features_piece_task}Features to characterize manufactured pieces (scenario 1) and characterize worker sessions (scenario 2).}
\begin{tabular}{ccp{5cm}p{7.5cm}}
\toprule
\bf Scenario & \bf Number & \multicolumn{1}{c}{\bf Name} & \multicolumn{1}{c}{\bf Description}\\ \hline

\multirow{7}{*}{1}

& \multirow{1}{*}{1} & \multirow{1}{*}{Piece \textsc{id}} & Identifier of the manufactured piece.\\

& \multirow{1}{*}{2} & \multirow{1}{*}{Input instant} & Instant when the piece is fed into the weight sensor.\\

& \multirow{1}{*}{3} & \multirow{1}{*}{Output delay} & Instant when the piece leaves the conveyor belt.\\

& \multirow{1}{*}{4} & \multirow{1}{*}{Time between pieces} & Time since one piece leaves the conveyor belt until a new piece is fed into the weight sensor.\\

& \multirow{1}{*}{5} & \multirow{1}{*}{Type of piece} & Boolean value that indicates if the piece is valid (true) or invalid (false).\\

\hline

\multirow{25}{*}{2}

& \multirow{1}{*}{1} & \multirow{1}{*}{ Session \textsc{id}} & Identifier of the session.\\

& \multirow{1}{*}{2} & \multirow{1}{*}{Input instants} & Vector of input instants of the pieces.\\

& \multirow{1}{*}{3} & \multirow{1}{*}{Output delays} & Vector of output delays of the pieces.\\ 

& \multirow{1}{*}{4} & \multirow{1}{*}{Number of incidences} & Number of incidences of the manufacturing machine.\\ 

& \multirow{1}{*}{5} & \multirow{1}{*}{Number of invalid pieces} & Number of invalid pieces according to video analysis.\\ 

& \multirow{1}{*}{6} & \multirow{1}{*}{Number of valid pieces} & Number of valid pieces according to video analysis.\\

& \multirow{1}{*}{7} & \multirow{1}{*}{Number of directly placed pieces} & Number of pieces that are directly taken from the warehouse tray and placed into the weight sensor.\\

& \multirow{1}{*}{8} & Number of pieces collected from the warehouse tray
& Total number of pieces collected from the warehouse tray.\\

& \multirow{1}{*}{9} & \multirow{1}{*}{Number of pieces taken to the buffer} & Total number of pieces collected from the warehouse tray and placed temporarily in a buffer before taking them to the weight sensor.\\

& \multirow{1}{*}{10} & \multirow{1}{*}{Number of reloads} & Number of workers reloads of material, either from the warehouse tray or from the buffer.\\

& \multirow{1}{*}{11} & \multirow{1}{*}{Number of mobile assistant reboots} & Number of mobile assistant reboots due to incidences.\\

& \multirow{1}{*}{12} & \multirow{1}{*}{Pieces' types} & Boolean vector whose dimension is the number of produced pieces: true represents a valid piece, false an invalid piece.\\

& \multirow{1}{*}{13} & \multirow{1}{*}{Time between pieces} & Vector of times between successive pieces (a piece leaving the conveyor belt and the next input piece being placed into the weight sensor).\\

& 14 & Time between valid pieces & Vector of times between successive valid pieces.\\

& \multirow{1}{*}{15} & \multirow{1}{*}{Total time} & Time length of the session.\\

\bottomrule
\end{tabular}
\end{table*}

\subsection{Data processing, feature engineering, analysis \& selection}
\label{sec:data_processing_results}

Scenario 1 focuses on piece profiling. It used \num{283} registered entries from the Elasticsearch database with the five features in Table \ref{tab:features_piece_task}. The target variable divides these entries into \num{165} pieces manufactured by expert users and \num{118} pieces manufactured by inexpert users. 

For scenario 2 on worker action profiling, data from \num{30} worker sessions was taken from the database, each of them characterized by the fifteen features in Table \ref{tab:features_piece_task} plus the averages and $Q1-Q3$ quartiles of features \#2, \#3, and \#12-\#14, yielding a total of 35 features per session including the session identifier. The target variable divides these worker sessions into \num{20} sessions of expert workers and \num{10} sessions of inexpert workers.

Table \ref{tab:correlations} shows the Pearson correlations between the features and the target variable whose absolute values exceed $\delta=0.2$.

\begin{table*}[!htbp]
\centering
\caption{\label{tab:correlations}Pearson correlations between features and target variable. Feature suffixes: (1) Average, (2) $Q_1$, (3) $Q_2$, (4) $Q_3$.}
\begin{threeparttable}
\begin{tabular}{ccccccccccc}
\toprule

\bf Scenario & \multicolumn{10}{c}{\bf Feature/correlation value} \\\midrule

\multirow{2}{*}{1} & \#2 & \#3 & \#5 \\

\cmidrule(lr){2-4}

& -0.56 & -0.57 & 0.30\\\midrule
 
\multirow{6}{*}{2} & \#2(1) & \#2(2) & \#2(3) & \#2(4) & \#3(1) & \#3(2) & \#3(3) & \#3(4) & \#5 & \#6\\ \cmidrule(lr){2-11}

& -0.86 & -0.84 & -0.87 & -0.86 & -0.88 & -0.35 & -0.89 & -0.87 & -0.84 & 0.47\\

\\
 
& \#7 & \#8 & \#9 & \#10 & \#11 & \#12(1) & \#12(2) & \#12(3) & \#13(1) & \#15\\ \cmidrule(lr){2-11}

& 0.45 & -0.82 & -0.99 & -0.50 & -0.32 & 0.76 & 0.83 & 0.47 & -0.72 & -0.91\\
 
\bottomrule
\end{tabular}
\end{threeparttable}
\end{table*}

The transformer wrapper \texttt{SelectFromModel}\footnote{Available at \url{https://scikit-learn.org/stable/modules/generated/sklearn.feature_selection.SelectFromModel.html}, June 2023.} was used for feature selection with an \textsc{rf} model with \num{50} estimators, while the rest of the parameters were kept by default. More in detail, \texttt{SelectFromModel} applies a univariate feature selection method using uni-variate statistical tests. This method allows discarding non-target-specific features, \textit{i.e.}, which contribute little to distinguish expert workers from inexpert ones. Finally, the selected features by both feature selection methods were:

\begin{description}

\item \textbf{Scenario 1}. Features \#2 and \#3 in Table \ref{tab:features_piece_task}.

\item \textbf{Scenario 2}. Average values of features \#2 and \#3, $Q1$, $Q2$, and $Q3$ values of feature \#3, and feature \#9 in Table \ref{tab:features_piece_task}.

\end{description}

For scenario 1, \num{283} samples were used with three features before feature selection (features \#2-\#4 in Table \ref{tab:features_piece_task}). As previously mentioned, features \#2 and \#3 in Table \ref{tab:features_piece_task} were used after feature selection, resulting in \num{849} data values. Scenario 2 had \num{30} samples with \num{29} features before feature selection (averages and $Q1-Q3$ quartiles of features \#2, \#3 and \#12-\#14, and features \#4-\#11 and \#15 in Table \ref{tab:features_piece_task}). After feature selection, \num{6} features were used (average values of features \#2 and \#3, $Q1-Q3$ quartiles of feature \#3, and feature \#9 in Table \ref{tab:features_piece_task}), resulting in \num{180} data values.

\subsection{Machine Learning classification}
\label{sec:classification_results}

The implementations of the models used are {\tt SVC}\footnote{Available at \url{https://scikit-learn.org/stable/modules/generated/sklearn.svm.SVC.html}, June 2023.} (with linear, polynomial, \textsc{rbf}, and sigmoid kernels), {\tt RandomForestClassifier}\footnote{Available at \url{https://scikit-learn.org/stable/modules/generated/sklearn.ensemble.RandomForestClassifier.html}, June 2023.}, and {\tt AdaBoostClassifier}\footnote{Available at \url{https://scikit-learn.org/stable/modules/generated/sklearn.ensemble.AdaBoostClassifier.html}, June 2023.}.

\begin{figure*}[!ht]
  \centering
  \subfloat[\centering Linear \textsc{svc}, scenario 1.]{{\includegraphics[width=4.5cm]{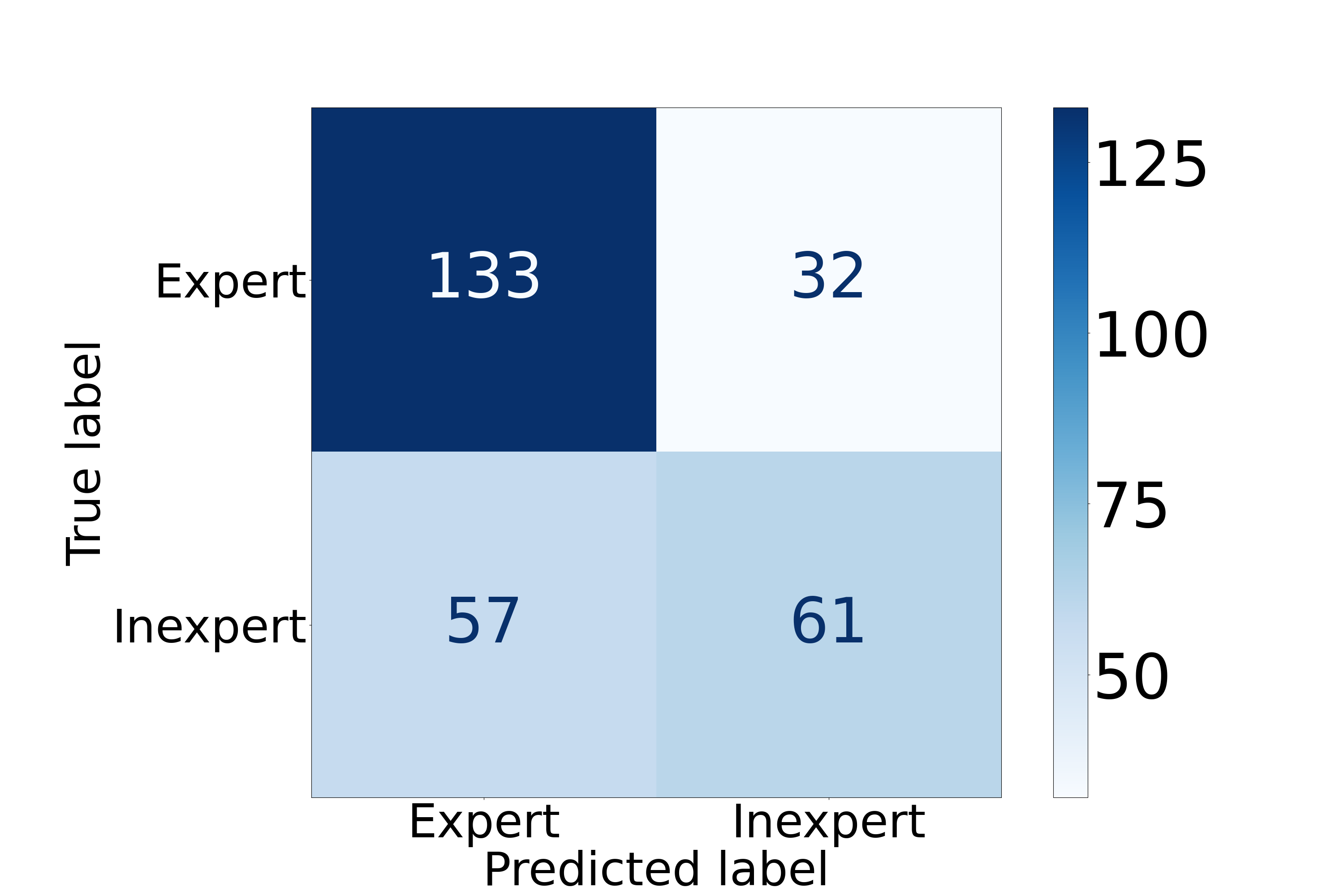}}}
  \qquad
  \subfloat[\centering \textsc{svc}-polynomial, scenario 1.]{{\includegraphics[width=4.5cm]{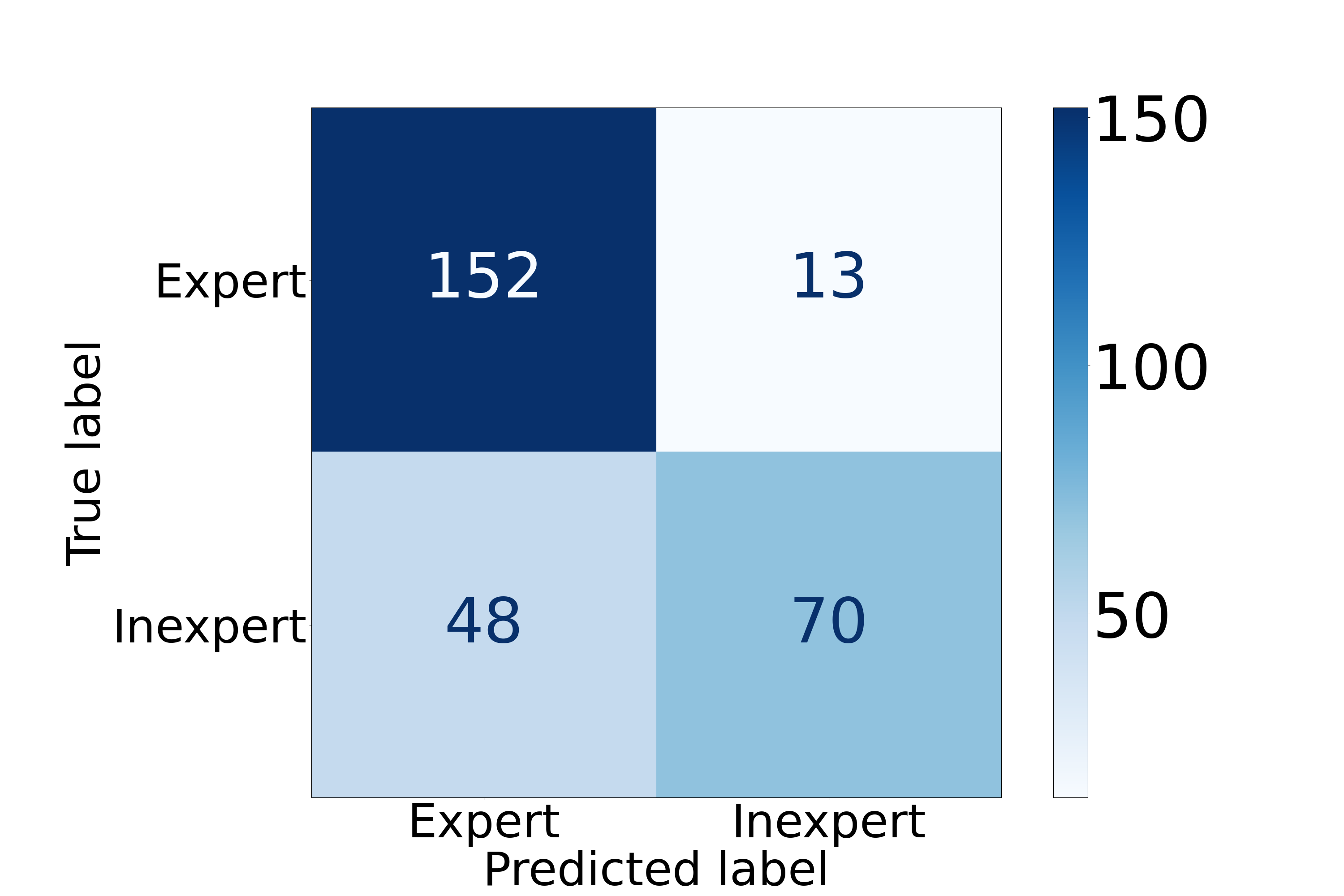}}}
  \qquad
\subfloat[\centering \textsc{svc-rbf}, scenario 1.]{{\includegraphics[width=4.5cm]{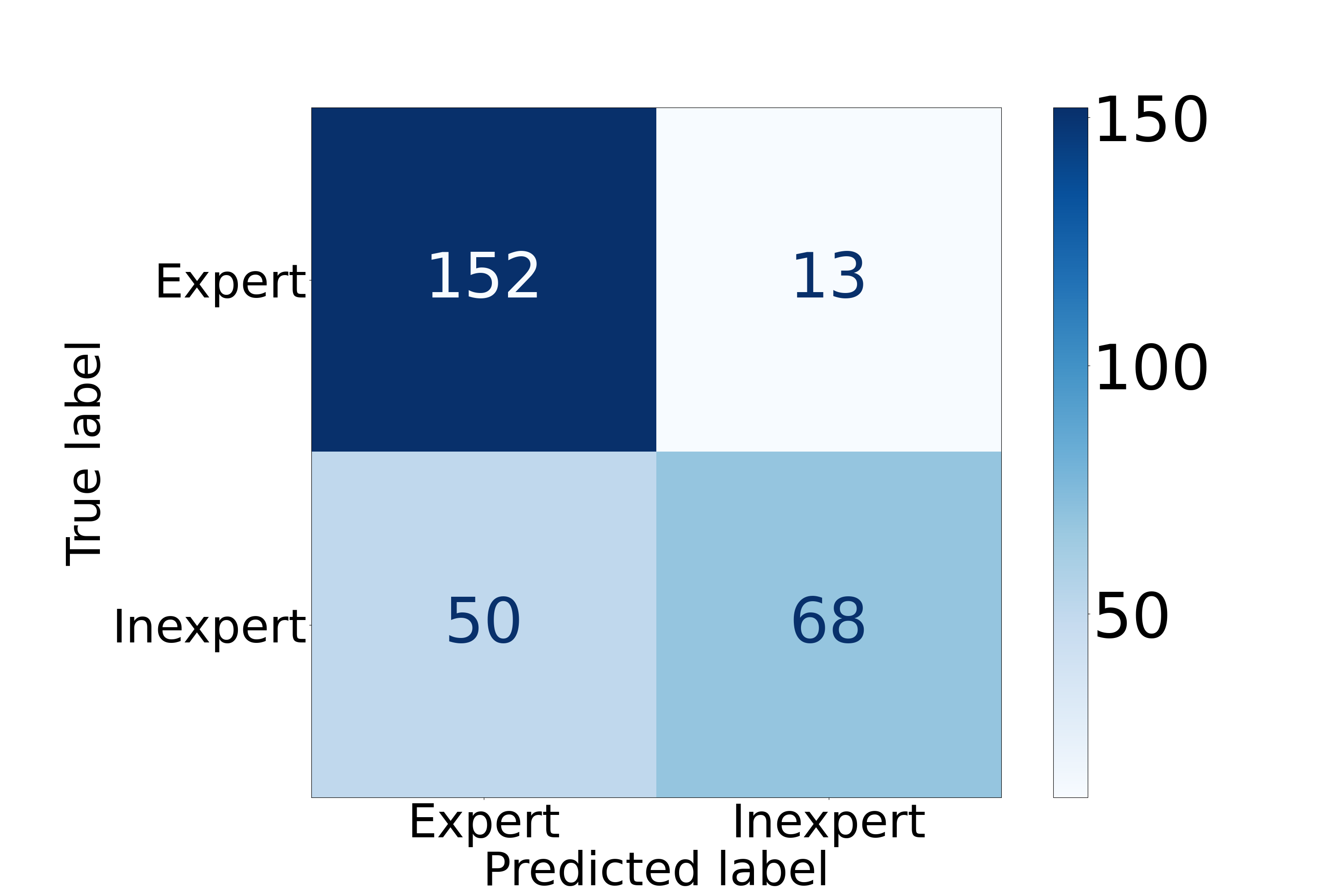}}}
  \qquad
   \subfloat[\centering \textsc{svc}-sigmoid, scenario 1.]{{\includegraphics[width=4.5cm]{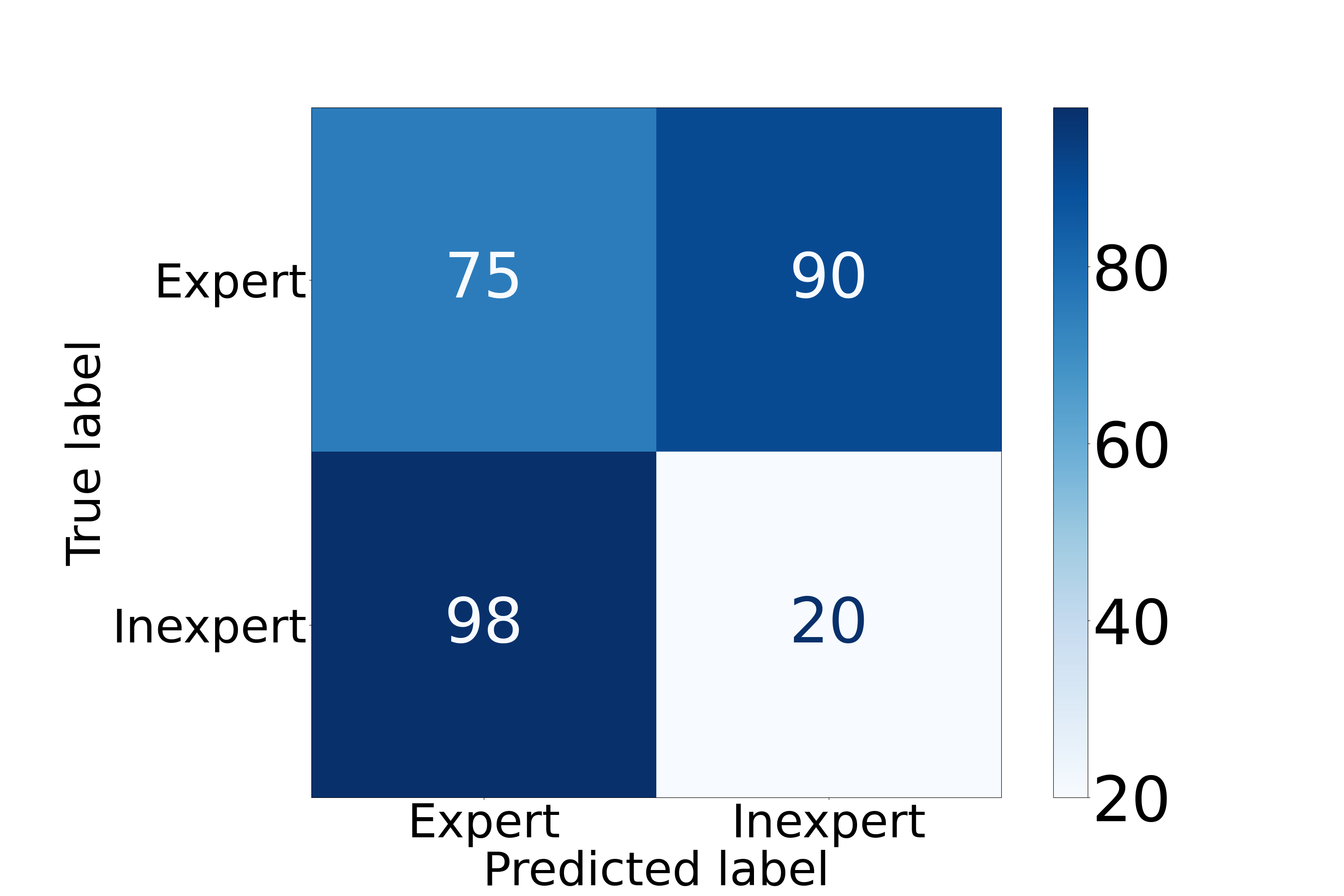}}}
  \qquad
  \subfloat[\centering \textsc{rf}, scenario 1.]{{\includegraphics[width=4.5cm]{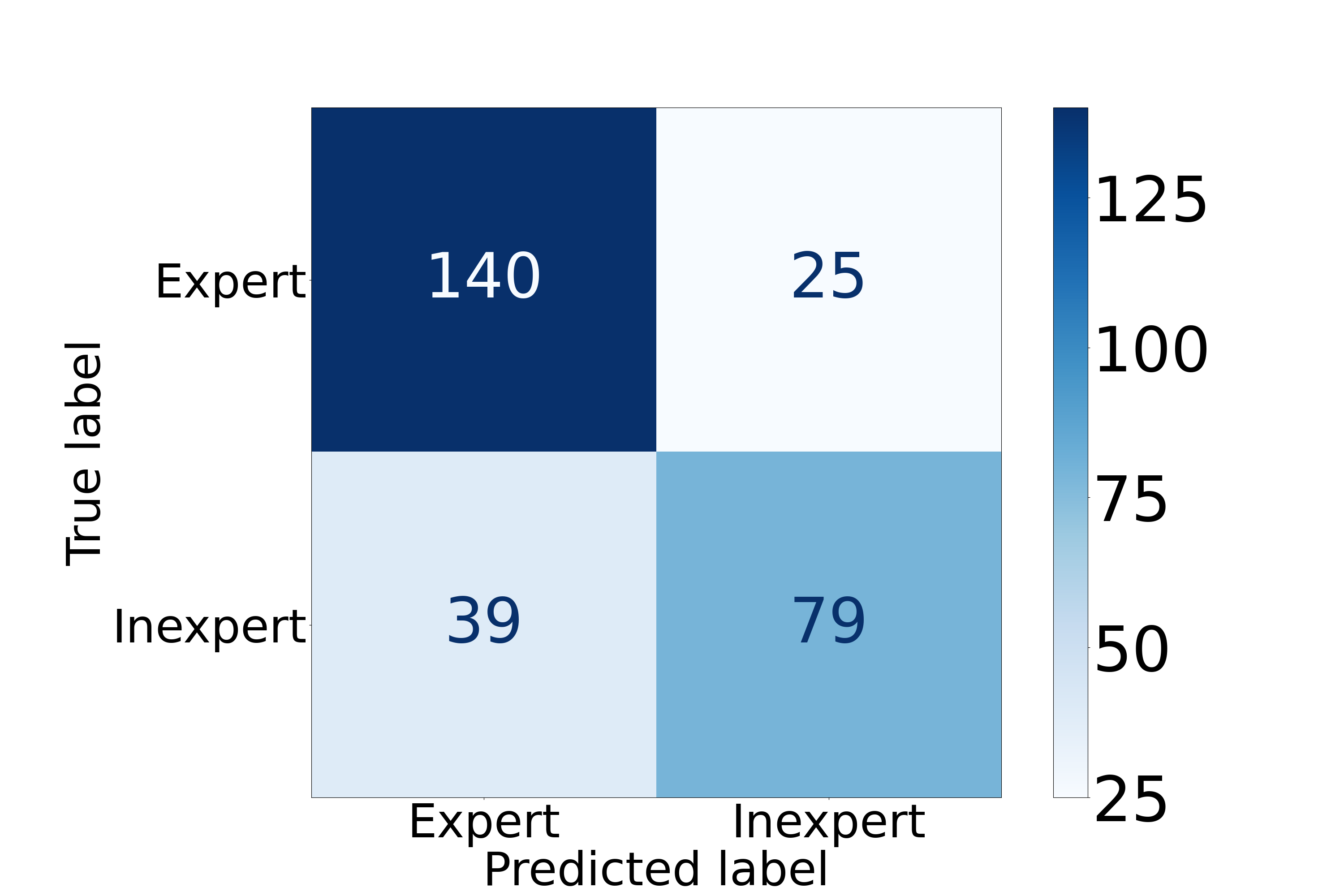}}}
  \qquad
  \subfloat[\centering \textsc{ab}, scenario 1.]{{\includegraphics[width=4.5cm]{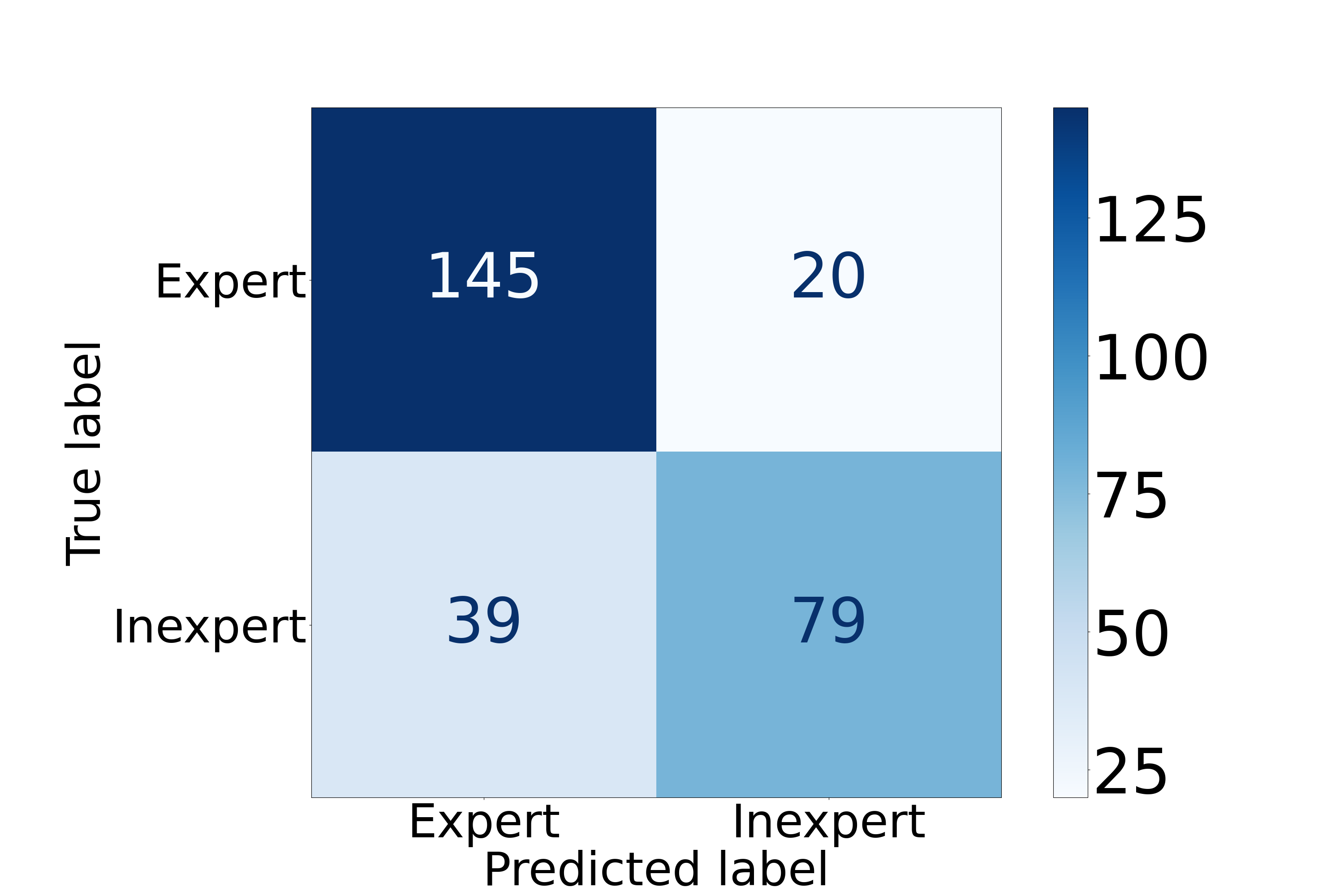}}}

  \subfloat[\centering Linear \textsc{svc}, scenario 2.]{{\includegraphics[width=4.5cm]{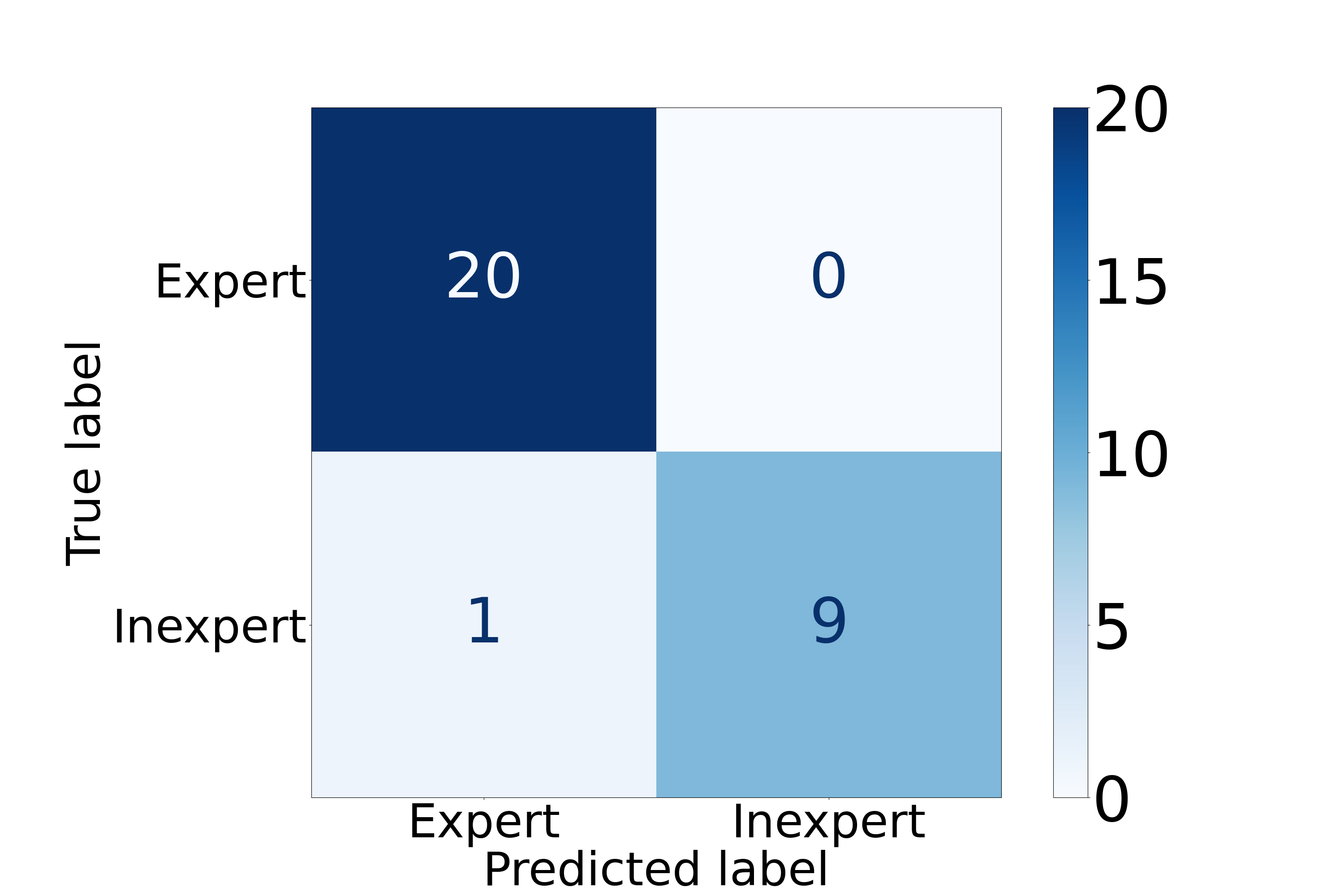}}}
  \qquad
    \subfloat[\centering \textsc{svc}-polynomial, scenario 2.]{{\includegraphics[width=4.5cm]{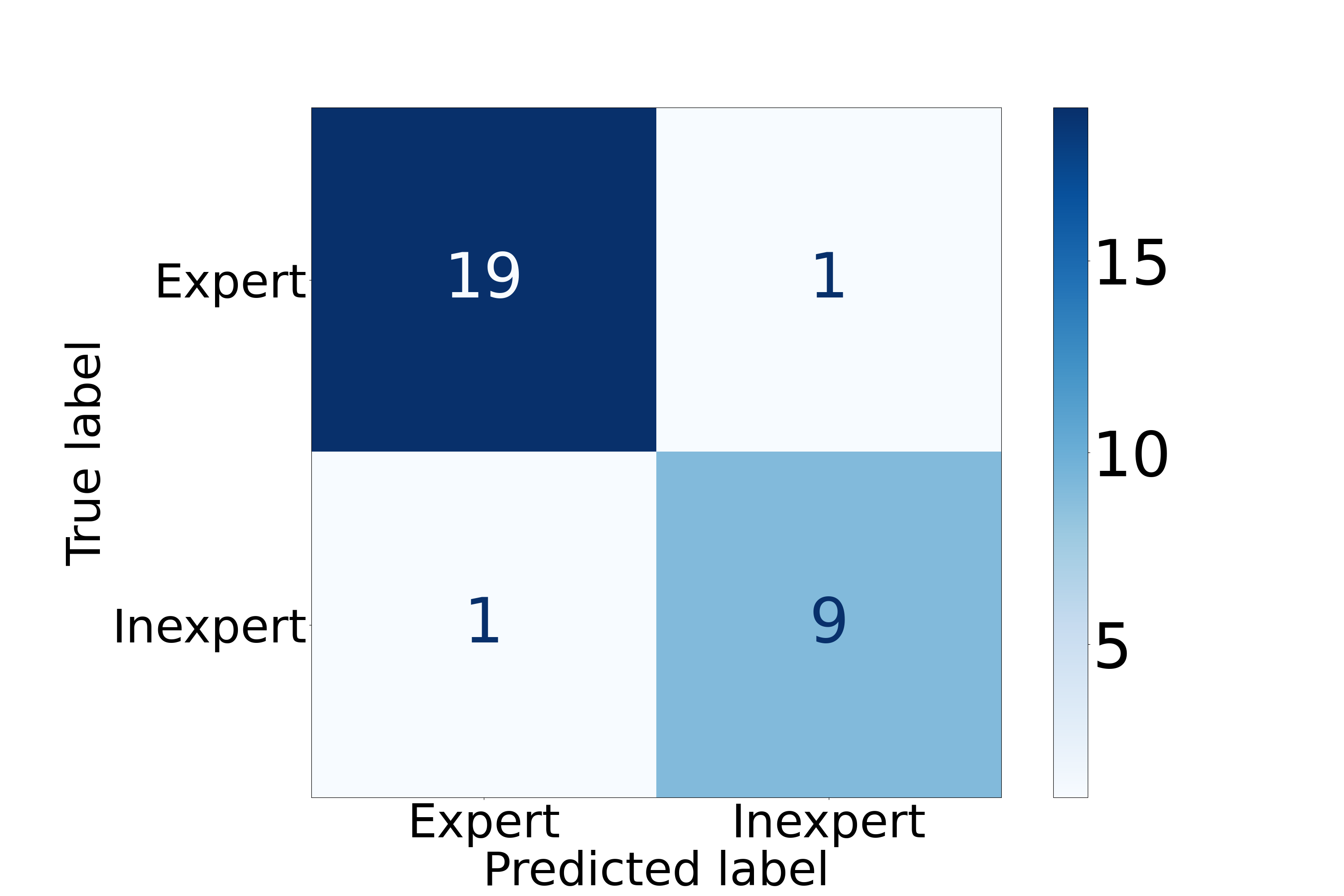}}}
  \qquad
  \subfloat[\centering \textsc{svc-rbf}, scenario 2.]{{\includegraphics[width=4.5cm]{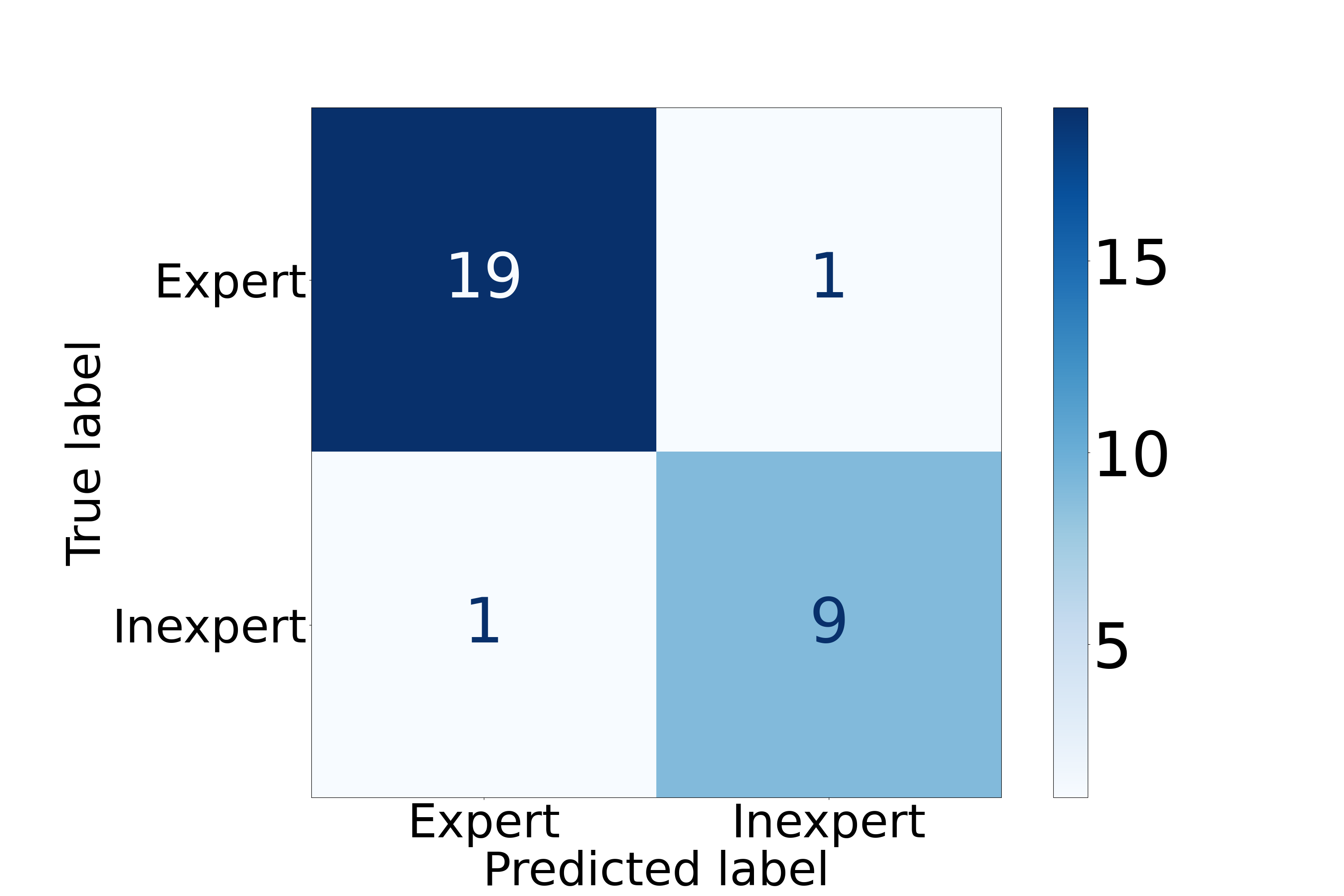}}}
  \qquad
   \subfloat[\centering \textsc{svc}-sigmoid, scenario 2.]{{\includegraphics[width=4.5cm]{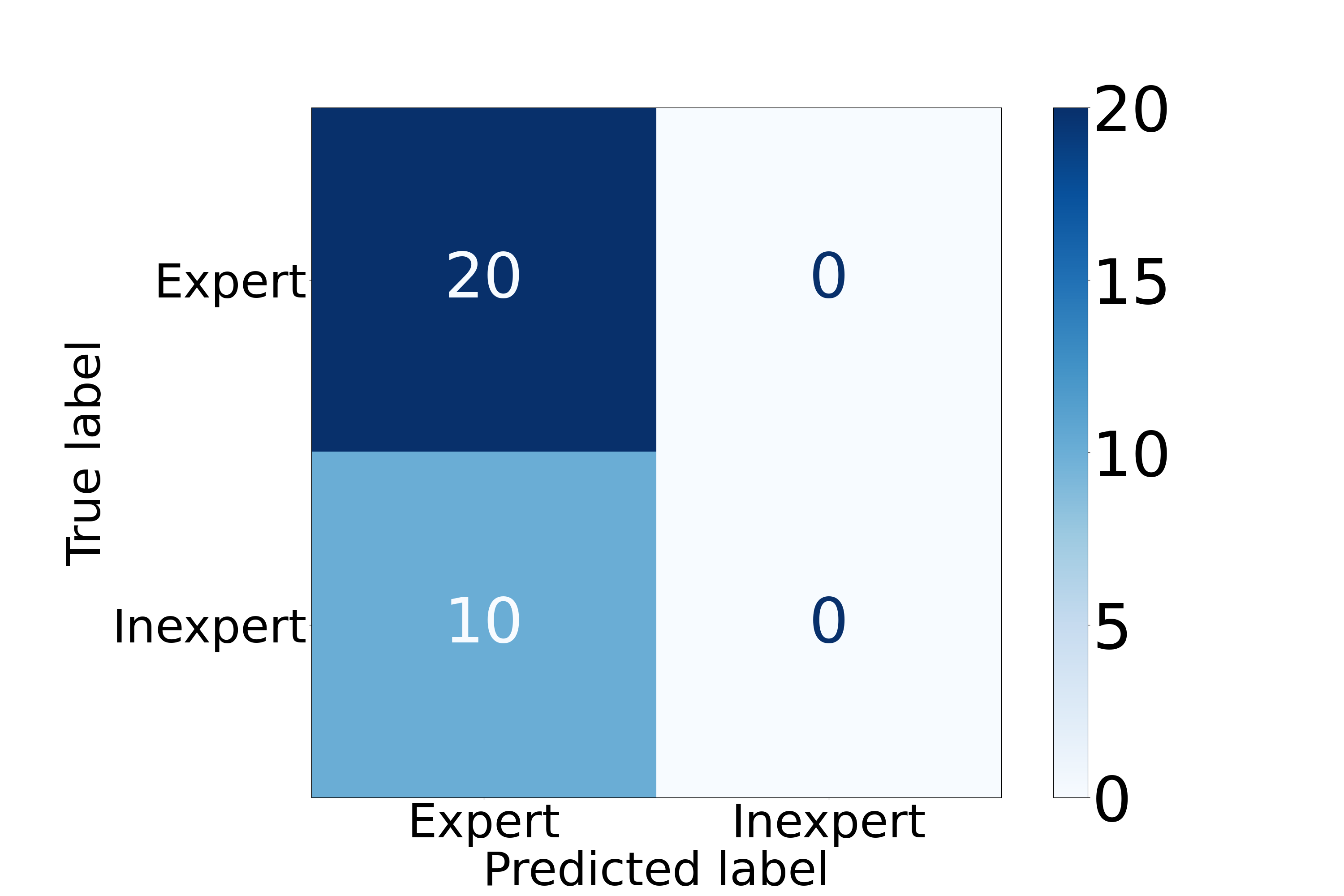}}}
  \qquad
  \subfloat[\centering \textsc{rf}, scenario 2.]{{\includegraphics[width=4.5cm]{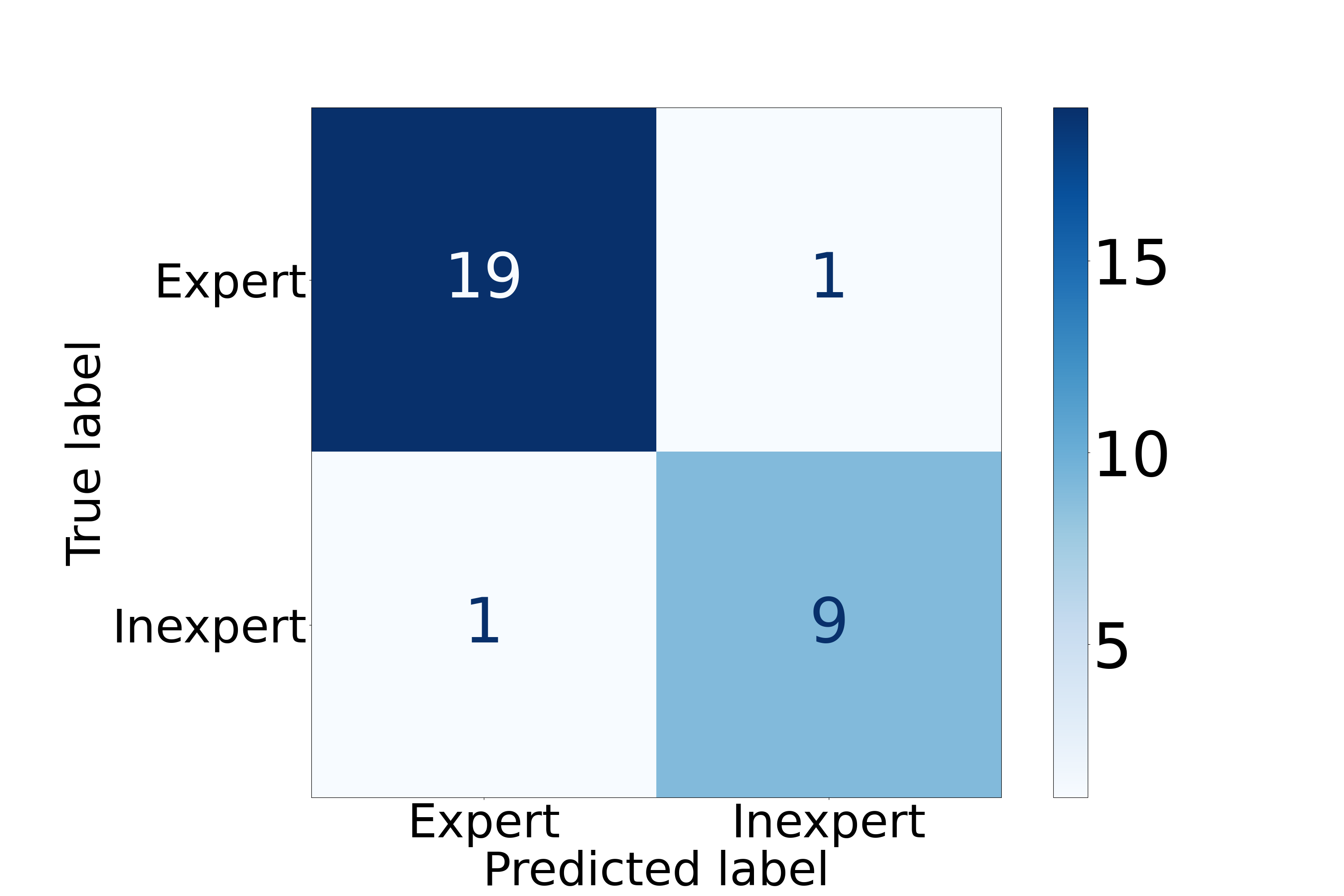}}}
  \qquad
  \subfloat[\centering \textsc{ab}, scenario 2.]{{\includegraphics[width=4.5cm]{fig5k.png}}}
  
  \caption{\label{fig:confusion_matrices}Confusion matrices of the different \textsc{ml} models for scenarios 1 and 2.}
  \qquad

\end{figure*}

Figure \ref{fig:confusion_matrices} shows the confusion matrices for the three \textsc{ml} models in scenarios 1 and 2.

\begin{figure*}[!ht]
  \centering

\end{figure*}

Tables \ref{tab:classification_results_macro} and \ref{tab:classification_results_micro}, respectively, show the macro and micro values of the results obtained using 10-fold cross-validation \cite{Berrar2019} in the two scenarios. This partitioning strategy was followed to avoid over-fitting since it typically reduces over- and underestimation. More in detail, in 10-fold cross-validation, the data set is divided ten times into 10 partitions, 9 for training and 1 for testing, without overlapped testing partitions. Finally, the overall performance metrics are averaged.

Differentiating between expert and inexpert workers at the piece level was more challenging. However, the results at the session level were very satisfactory, with nearly \SI{100}{\percent} performance metrics for all models. In this second scenario, even though the linear SVC and \textsc{ab} were the best performers, \textsc{rf} was selected with the explainability stage in mind as it is intrinsically interpretable. Even at the piece level, \textsc{rf} and \textsc{ab} could attain near-80 \% accuracy and precision.

Errors in scenario 1 may be produced by expert users that manufacture pieces as inexpert ones, and vice versa. The latter makes exceeding \SI{80}{\percent} accuracy difficult. Conversely, errors are minimized in scenario 2 thanks to the variety of input features that allow the classifier to differentiate between expert and inexpert workers, as reflected in the confusion matrices in Figure \ref{fig:confusion_matrices}. Accordingly, in scenario 1, the lowest recall value indicates that most of the confusion comes from incorrectly identifying experts as inexpert users (\SI{66.95}{\percent} and \SI{87.88}{\percent} recall for inexpert and expert workers, respectively, with the \textsc{ab} model). Note that \textit{F}-measure values for both expert and inexpert user detection, a more reliable measure, exceed \SI{70}{\percent} in scenario 1. Furthermore, in scenario 2, \textit{F}-measure values for the same purpose exceed \SI{95}{\percent}.

\begin{table*}[!htbp]
\centering
\caption{\label{tab:classification_results_macro}Classification results (macro values).}
\begin{tabular}{ccccccc}
\toprule
\bf {Scenario} & \bf {Classifier} & \bf {Accuracy} & \bf Precision & \bf Recall & \bf \textit{F}-measure & \bf Total time (s)\\
\midrule
\multirow{6}{*}{1} & 
Linear \textsc{svc} & 68.55\% & 67.80\% & 66.15\% & 66.37\% & 0.08\\
& \textsc{svc}-polynomial & 78.45\% & \bf 80.17\% & 75.72\% & 76.47\% & 0.04\\
& \textsc{svc-rbf} & 77.74\% & 79.60\% & 74.87\% & 75.59\% & 0.03\\
& \textsc{svc}-sigmoid & 33.57\% & 30.77\% & 31.20\% & 30.96\% & 0.04\\
& \textsc{rf} & 77.39\% & 77.09\% & 75.90\% & 76.28\% & 1.72\\
& \textsc{ab} & \bf 79.15\% &  79.30\% & \bf 77.41\% & \bf 77.95\% & 0.52\\
\midrule

\multirow{6}{*}{2} & 
Linear \textsc{svc} & \bf 96.67\% & \bf 97.62\% & 95.00\% & 96.15\% & 0.03\\
& \textsc{svc}-polynomial & 93.33\% & 92.50\% & 92.50\% & 92.50\% & 0.02\\
& \textsc{svc-rbf} & 93.33\% & 92.50\% & 92.50\% & 92.50\% & 0.02\\
& \textsc{svc}-sigmoid & 66.67\% & 33.33\% & 50.00\% & 40.00\% & 0.02\\
& \textsc{rf} &93.33\% & 92.50\% & 92.50\% & 92.50\% & 1.70\\
& \textsc{ab} & \bf 96.67\% & 95.45\% & \bf 97.50\% & \bf 96.34\% & 0.03\\
\bottomrule
\end{tabular}
\end{table*}

\begin{table*}[!htbp]
\centering
\caption{\label{tab:classification_results_micro}Classification results (micro values per class).}
\begin{tabular}{cccccccc}
\toprule
\bf {Scenario} & \bf {Classifier} & \multicolumn{2}{c}{\bf Precision} & \multicolumn{2}{c}{\bf Recall} & \multicolumn{2}{c}{\bf \textit{F}-measure}\\

\cmidrule(lr){5-6}
\cmidrule(lr){7-8}
& & Inexpert & Expert & Inexpert & Expert & Inexpert & Expert\\
\midrule
\multirow{6}{*}{1} & 
Linear \textsc{svc} & 65.59\% & 70.00\% & 51.69\% & 80.61\% & 57.82\% & 74.93\%\\
& \textsc{svc}-polynomial & \bf84.34\% & 76.00\% & 59.32\% & \bf 92.12\% & 69.65\% & \bf83.29\%\\
& \textsc{svc-rbf} & 83.95\% & 75.25\% & 57.63\% & \bf 92.12\% & 68.34\% & 82.83\%\\
& \textsc{svc}-sigmoid & 18.18\% & 43.35\% & 16.95\% & 45.45\% & 17.54\% & 44.38\%\\
& \textsc{rf} & 75.96\% & 78.21\% & 66.95\% & 84.85\% & 71.17\% & 81.40\%\\
& \textsc{ab} & 79.80\% & \bf 78.80\% & \bf 66.95\% & 87.88\% & \bf 72.81\% & 83.09\%\\
\midrule

\multirow{6}{*}{2} & 
Linear \textsc{svc} & \bf 100.00\% & 95.24\% & 90.00\% & \bf 100.00\% & 94.74\% & \bf 97.56\%\\
& \textsc{svc}-polynomial & 90.00\% & 95.00\% & 90.00\% & 95.00\% & 90.00\% & 95.00\%\\
& \textsc{svc-rbf} & 90.00\% & 95.00\% & 90.00\% & 95.00\% & 90.00\% & 95.00\%\\
& \textsc{svc}-sigmoid & 0.00\% & 66.67\% & 0.00\% & \bf 100.00\% & 0.00\% & 80.00\%\\
& \textsc{rf} & 90.00\% & 95.00\% & 90.00\% & 95.00\% & 90.00\% & 95.00\%\\
& \textsc{ab} & 90.91\% & \bf 100.00\% & \bf 100.00\% & 95.00\% & \bf 95.24\% & 97.44\%\\
\bottomrule
\end{tabular}
\end{table*}

At this point, comparing the classification performance with previous works on related problems is illustrative. The solution by Lather \textit{et al.} \cite{Lather2019} to assess worker performance had \SI{79.90}{\percent} accuracy, \SI{87.00}{\percent} precision, and \SI{85.00}{\percent} recall with the \textsc{rf} classifier (\SI{13.43}{\percent}, \SI{5.50}{\percent} and \SI{7.50}{\percent} less than in our problem, respectively). Forkan \textit{et al.} \cite{Forkan2019} attained \SI{71.57}{\percent} accuracy (\SI{21.76}{\percent} less) for worker activity recognition with the same \textsc{ml} model using hand accelerometer sensors. Al \textit{et al.} (2019) \cite{AlJassmi2019} obtained \SI{66.11}{\percent} accuracy for worker productivity prediction using psychological signals from wearables also with an \textsc{rf} classifier (\SI{27.22}{\percent} less than in our problem). Patalas-Maliszewska and Halikowsk \cite{Patalas-Maliszewska2020} differentiated tasks by correctness with \SI{73.15}{\percent} accuracy using the deep learning \textsc{yolo} object detection module (\SI{20.18}{\percent} less than in our problem). Logically, these works' experimental settings differ, but the general performance references are comparable to our worst-case scenario.

Summing up, achieved \textsc{ml} accuracy is satisfactory for worker performance characterization, as supported by the comparison with related works from the literature. Even though other industrial \textsc{ml} problems (\textit{e.g.}, on machine characterization) may require more stringent performance levels, our final goal is to extract valuable insights by applying explainability techniques, and the classification accuracy level is subordinated to that goal.

\subsection{Worker \textsc{kpi}s}
\label{sec:worker_kpis_results}

As previously said, no standard Worker/Operator 4.0
\textsc{kpi}s yet exist to characterize behaviors such as over- and under-performance of the workers. Therefore, to support the explainability of the classification decisions of scenarios 1 and 2, average, $Q1$, and $Q3$ values were calculated for the variables in Table \ref{tab:kpis} by following two approaches: 

\begin{description}

\item \textbf{Intra-worker performance analysis}. Daily values of the \textsc{kpi}s were aggregated weekly to compare a worker's performance with himself/herself for the previous 7 days.

\item \textbf{Inter-worker performance analysis}. Daily values of the \textsc{kpi}s were used to compare different workers.

\end{description}

Table \ref{tab:kpis} also shows the triggers we chose to differentiate between expert and inexpert workers in the cases when $Q_1<$average$<Q_3$, thus focusing on those cases with significant statistical deviation.

\begin{table*}[!htbp]
\centering
\caption{\label{tab:kpis}Worker \textsc{kpi}s.}
\begin{tabular}{ccccp{5cm}}
\toprule
\bf Number & \multicolumn{1}{c}{\bf Expert trigger}&\multicolumn{1}{c}{\bf Name}& \multicolumn{1}{c}{\bf Inexpert trigger}\\ \hline

\multirow{1}{*}{1} & $Q1>N_{inc}$ & \multirow{1}{*}{$N_{inc}$: number of incidences} & $N_{inc}>Q3$\\

\multirow{1}{*}{2} & $Q1>N_{inv}$ &\multirow{1}{*}{$N_{inv}$: number of invalid pieces} & $N_{inv}>Q3$\\

\multirow{1}{*}{3} & $Q3<N_{val}$ & \multirow{1}{*}{$N_{val}$: number of valid pieces} & $N_{val}<Q1$\\

\multirow{1}{*}{4} & $Q3<N_{task}$ &\multirow{1}{*}{$N_{task}$: number of tasks} & $N_{task}<Q1$\\

\multirow{1}{*}{5} & $Q1>T_{val}$ & \multirow{1}{*}{$T_{val}$: time between valid pieces} & $T_{val}>Q3$\\

\multirow{1}{*}{6} & $Q1>T_{total}$ & \multirow{1}{*}{$T_{total}$: total time} & $T_{total}>Q3$\\

\bottomrule
\end{tabular}
\end{table*}

\subsection{Explainability}
\label{sec:explainability_results}

The \textsc{lime} library\footnote{Available at \url{https://github.com/marcotcr/lime}, June 2023.} was used to obtain the most relevant features involved in the predictions at the sample level. According to the average relevance scores obtained for identifying expert workers, the features can be ordered as follows in decreasing relevance for the two scenarios:
\begin{description}
\item \textbf{Scenario 1}: \#3, \#2.
\item \textbf{Scenario 2}: \#9, \#3(Q3), \#3(Q2), \#3(avg), \#3(Q1) \#2(avg).
\end{description}

The most relevant features for scenario \#1 to identify expert workers are coherent with that purpose since they contribute to increasing piece throughput. However, in scenario \#2, they also provide insights about the worker actions that improve performance: feature \#9 allowed identifying a particular behavior of expert workers, of keeping the buffer full to feed the weight sensor as quickly as possible. The latter seemed logical a posterior, but it was not included in the step-by-step mobile assistant instructions in the manual. 

From the explainability analysis and the \textsc{kpi} analysis, it is possible to provide textual and visual descriptions to the plant managers and workers about their performance. These descriptions include information about the most relevant features and, most importantly, insights about worker behavior based on the values of the features. 

Next, we present examples of automatic textual reports based on explainability templates.

Regarding scenario 1, which focuses on the manufactured piece, the first statement of the textual report highlights the relevance of the output delay by checking the decision thresholds of relevant feature \#3 of the \textsc{rf} classification model (see Table \ref{tab:features_piece_task}). Then, in the second statement, the predicted piece category is explained in terms of confidence, critical features involved, and classification output (expert worker in this case).

\begin{enumerate}

\item For \textcolor{cyan}{piece} 2 produced by \textcolor{cyan}{worker} 7, the feature with \textcolor{cyan}{significant value} is:

\begin{itemize}
  \item \textcolor{cyan}{Output delay}: 26.00s < \textcolor{cyan}{output delay} $\leqslant$ 34.00 s
\end{itemize}

\item \textcolor{cyan}{Prediction} (\SI{78}{\percent} confidence): \textcolor{cyan}{the time reflected in the output delay} indicates that \textcolor{cyan}{piece} 2 has been produced by an \textcolor{cyan}{expert worker}.

\end{enumerate}

Similarly, regarding scenario 2, which focuses on the task performed, for the first statement of the textual report, under-performance and over-performance are also decided by checking the decision thresholds of relevant features, in this case, \#3 and \#9 of the \textsc{rf} algorithm. Then, the second statement explains the prediction by following the same method as in scenario 1. Intra-worker performance in statement 3 is described using \textsc{kpi}s \#1 and \#3 in Table \ref{tab:kpis}. Inter-worker performance in statement 4 is described using \textsc{kpi}s \#4 and \#6 in Table \ref{tab:kpis}. The report ends regarding worker skills and a general observation derived from feature \#3.

\begin{enumerate}

\item For \textcolor{cyan}{task} 10 performed by \textcolor{cyan}{worker} 7, features with \textcolor{cyan}{significant values} are:

\begin{itemize}
  \item \textcolor{cyan}{Under-performance}: \textcolor{cyan}{output delays average} > 44.61 s
  \item \textcolor{cyan}{Over-performance}: \textcolor{cyan}{number of pieces taken to the buffer} > 11.75 units
\end{itemize}
 
\item \textcolor{cyan}{Prediction} (\SI{91}{\percent} confidence): the \textcolor{cyan}{high number of pieces taken to the buffer} despite the \textcolor{cyan}{manufacturing time} indicates that \textcolor{cyan}{task} 10 has been performed by an \textcolor{cyan}{expert worker}.

\item \textcolor{cyan}{Intra-worker performance}: compared to the last 7 days, even though the \textcolor{cyan}{number of incidences} is \textcolor{cyan}{high}, \textcolor{cyan}{many valid pieces} have been created.

\item \textcolor{cyan}{Inter-worker performance}: compared to other workers, the \textcolor{cyan}{number of daily tasks} and the \textcolor{cyan}{manufacturing time} are \textcolor{cyan}{higher}.

\item Summing up, \textcolor{cyan}{worker} 7 exhibits \textcolor{cyan}{expert skills}, but the \textcolor{cyan}{manufacturing time} needs to improve.

\end{enumerate}

\begin{figure*}[!htbp]
\centering
\includegraphics[scale=0.15]{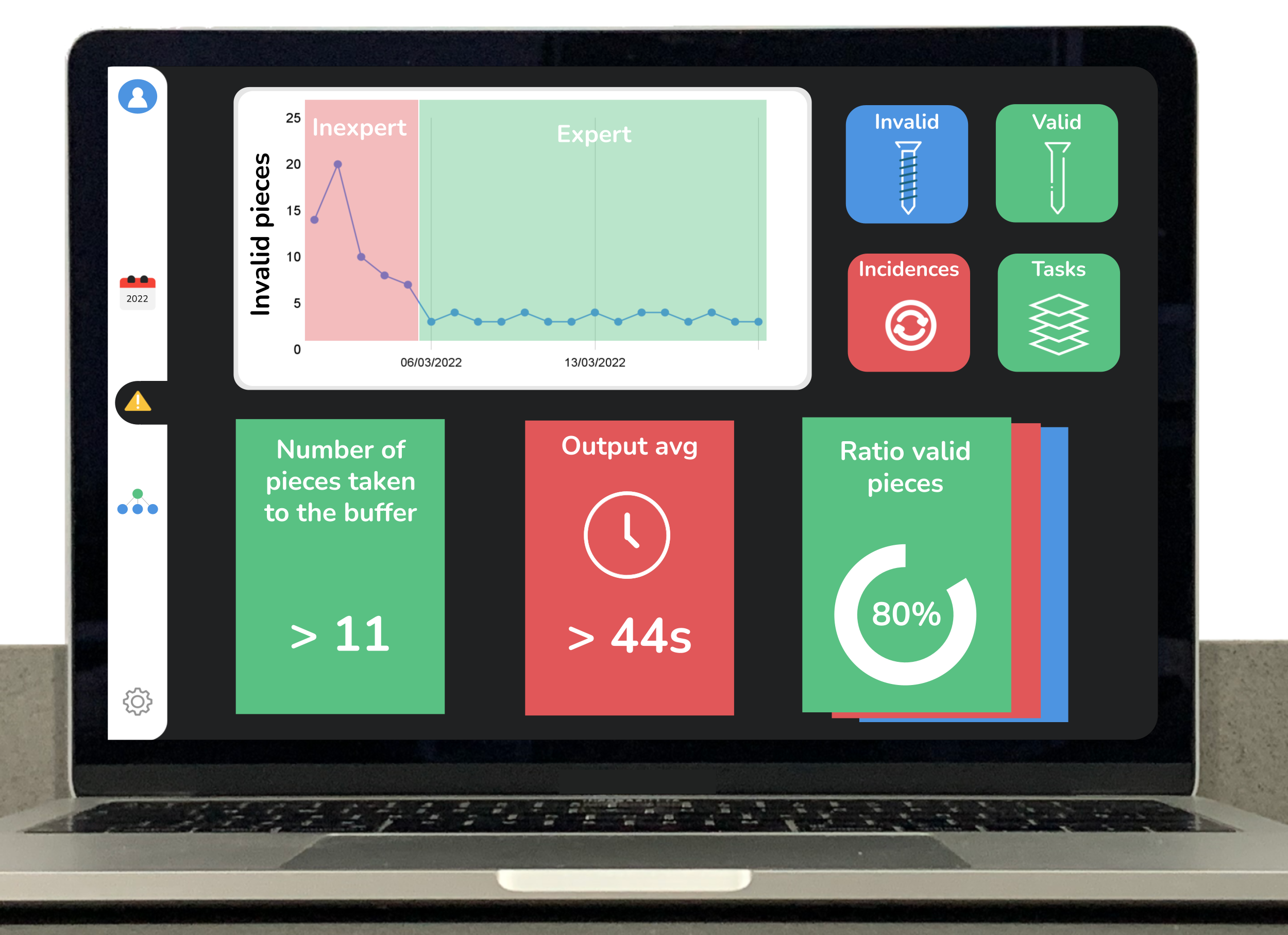}
\caption{\label{fig:dashboard}Explainability dashboard.}
\end{figure*}

Figure \ref{fig:dashboard} shows a view of the explainability dashboard from the points of view of piece validity and task.

In the dashboard, \textsc{kpi}-related boxes are displayed at the top right, and feature-related ones at the bottom. Boxes in blue correspond to neutral performance, and those in green and red to over- and under-performance, respectively. The graph at the top left shows successive classification results. In this regard, the user exhibited inexpert behavior for some time before acquiring expert skills with practice, yet with many incidences (\textsc{kpi} 1 in Table \ref{tab:kpis}). The number of invalid pieces (\textsc{kpi} 2 in Table \ref{tab:kpis}) obtained is acceptable, and the number of valid ones (\textsc{kpi} 3 in Table \ref{tab:kpis}) and tasks (sessions) performed (\textsc{kpi} 4 in Table \ref{tab:kpis}) are high. The number of pieces taken to the buffer and the average output time (features \#9 and \#3(1) of scenario 2 in Table \ref{tab:features_piece_task}) have been added for their direct and inverse impact on performance, respectively. The ratio of valid pieces in the bottom right is calculated as each task's average of \#6/(\#5+\#6). This box is green because the ratio exceeds \SI{66.67}{\percent}.

\section{Conclusions}
\label{sec:conclusions}

Most existing research on Industry 4.0 focuses on machines owing
to the extensive automation of manufacturing plants nowadays rather than
on the efficiency and productivity of workers. However, workers are central to the Worker/Operator 4.0 paradigm as part of Industry 4.0. In this context, \textsc{ai}, and \textsc{ml} techniques, in particular, are powerful tools for analyzing workers' performance in digitized workflows.

This work contributes to this field with a first proposal for applying explainability techniques to extract insights from \textsc{ml} models that differentiate between inexpert and expert workers, which is interesting for developing automatic worker training tools. 

The approach has been tested in a quality control cobot workstation with a mobile assistant interface. Departing from good skill level classification metrics nearing \SI{90}{\percent}, relevant features of an \textsc{rf} model and production \textsc{kpi}s (which were formulated for this work due to the lack of standardized indicators) were selected for highly informative textual and graphical explainability reports. Automatic explainability results allowed identifying productive actions of expert workers that were not included in the manual of the mobile assistant.

The proposed methodology can be enriched with additional features (for example, comments on the manufactured pieces) entered into the system by a supervisor, that is, in a human-in-the-loop scenario. In fact, at the cost of some manual effort, the methodology can be applied in traditional manufacturing stations with lower sensing just for automatic explainability purposes. Moreover, it can also lay the foundations for the automatic explanation of actions related to worker health problems (\textit{e.g.}, burnout syndrome \cite{LASTOVKOVA2018}).

Regarding how the proposal adds practical value to existing relevant industrial use cases, it can be exploited to overcome some barriers identified by P. Kumar \textit{et al.} (2021) \cite{Kumar2021}: (\textit{i}) workforce lacking adequate skills and job disruptions (codes \textsc{bri}4 and \textsc{bri}7, respectively) since with the system it is possible to infer knowledge from expert workers for inexpert ones to address future needs during the manufacturing process, (\textit{ii}) challenges in data management (\textsc{bri}8), since the solution can handle data from multiple sources (\textit{i.e.}, sensors, machines, etc.) and has the potential to extract information about the quality of the manufacturing products, and (\textit{iii}) resistance to change (\textsc{bri}10) to reduce workers unwillingness to the adoption of new technologies.

In future work, we plan to apply this approach to other workstations across different industrial workflows, including additional data such as real-time worker positioning. Data gathered from body sensors will also be exploited in an online \textsc{ml} pipeline to perform a complete and automatically explainable analysis of workers' actions. For example, to obtain insights into the effects of workers' postures.

\section*{Acknowledgements}

This work was partially supported by Xunta de Galicia grants ED481B-2021-118, ED481B-2022-093, and ED431C 2022/04, Spain; Ministerio de Ciencia e Innovación grant PID2020-116329GB-C21, Spain; and project SARATROF IN852B 2021/16, funded by Xunta de Galicia and co-funded by European Union Feder Galicia 2014-2020.

\bibliographystyle{IEEEtran}
\bibliography{mybibfile}

\end{document}